\documentclass[11pt]{article}

\usepackage[preprint]{acl}
\usepackage{pdflscape} % landscape mode
\usepackage[edges]{forest}
\usepackage{times}
\usepackage{latexsym}
\usepackage{amssymb}
\usepackage{booktabs}
\usepackage{float}
\usepackage{cuted}
\usepackage{caption}
\usepackage[T1]{fontenc}
\usepackage[utf8]{inputenc}

\usepackage{microtype}

\usepackage{inconsolata}

\usepackage{graphicx}

\usepackage[most]{tcolorbox}
\usepackage{xcolor}
\tcbuselibrary{listings, skins, breakable}
\usepackage{cuted}
\usepackage{placeins}
\usepackage{graphicx} 
\usepackage{makecell} 

\usepackage{tcolorbox}
\tcbset{
	myprompt/.style={
		enhanced,
        breakable, 
		colback=blue!3!white,        % background color
		colframe=cyan!85!white,      % border color
		fonttitle=\bfseries,
		title=Prompt~\thetcbcounter, % automatic numbering
		coltitle=black,
		sharp corners,
		boxrule=0.8pt,
		left=5pt, right=5pt, top=5pt, bottom=5pt,
		breakable,                   % allow page breaks
		before skip=10pt,
		after skip=10pt,
		listing only,
		listing options={
			basicstyle=\ttfamily\small,
			breaklines=true,
			columns=fullflexible,
			keepspaces=true,   % <--- preserve spaces
		},
	}
}
\newtcolorbox{promptbox}[1][]{%
    breakable,          % <--- THIS IS THE CRITICAL FIX
    colback=white,      % Adjust to your existing colors
    colframe=blue!50!black, 
    title={#1}
}

\newtcolorbox{examplebox}[2][]{
  enhanced,
  breakable,
  width=\textwidth,
  colback=gray!3!white,
  colframe=gray!60!black,
  title={Example #2},
  fonttitle=\bfseries,
  boxrule=0.8pt,
  rounded corners,
  sharp corners=south,
  before skip=10pt,
  after skip=10pt,
  attach title to upper,
  varwidth boxed title,
  left=8pt, right=6pt, top=6pt, bottom=6pt,
  sidebyside,
  sidebyside align=top, % Aligns the top of both columns
  lefthand width=0.35\textwidth, % Sets the width of the left column (for the image)
  #1
}

\newcommand{\rotHeader}[1]{\makecell{\rlap{\rotatebox{90}{\thead{#1}}}}}

\title{\includegraphics[width=0.6cm]{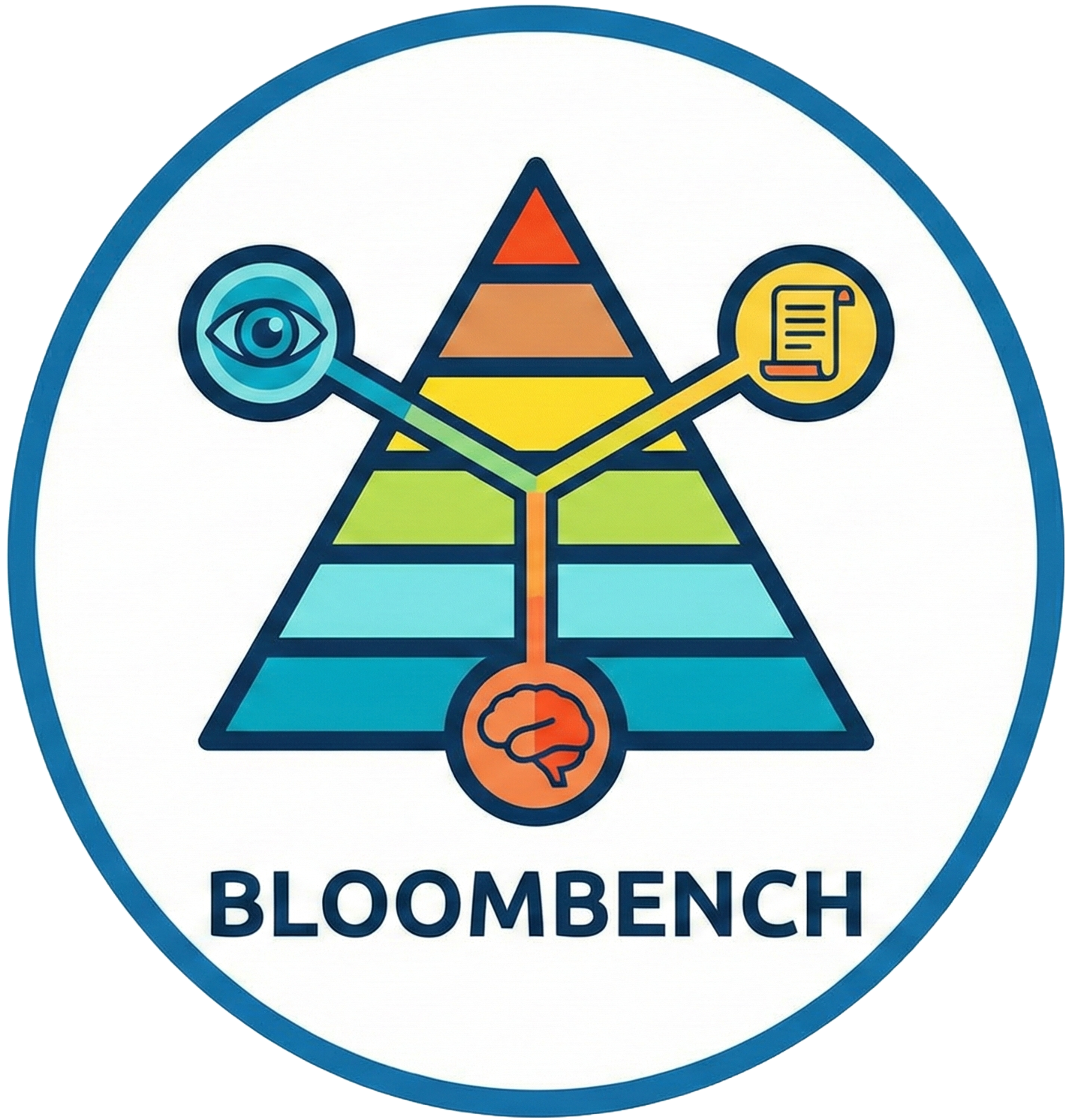}
Almieyar-Oryx-BloomBench: A Bilingual Multimodal Benchmark \\for Cognitively Informed Evaluation of Vision-Language Models
}

\author{
\textbf{Mohammad Mahdi Abootorabi\textsuperscript{\textdagger\textsection\textdaggerdbl,1}},
\textbf{Omid Ghahroodi\textsuperscript{\textdagger,1}},
\textbf{Anas Madkoor\textsuperscript{\textdagger}},\\
\textbf{Marzia Nouri\textsuperscript{\textdagger}},
\textbf{Doratossadat Dastgheib\textsuperscript{\textdagger}},
\textbf{Mohamed Hefeeda\textsuperscript{\textdagger}},
\textbf{Ehsaneddin Asgari\textsuperscript{\textdagger,*}}\\[0.4em]
\textsuperscript{\textdaggerdbl}University of British Columbia \quad
\textsuperscript{\textsection}Zuse School ELIZA\\
\textsuperscript{\textdagger}Qatar Computing Research Institute (QCRI), Hamad Bin Khalifa University\\[0.2em]
{\small \textsuperscript{1}Equal contribution \quad
\textsuperscript{*}Corresponding author: \href{mailto:easgari@hbku.edu.qa}{easgari@hbku.edu.qa}}
}

\usepackage{titlesec}

\titlespacing{\section}{0pt}{0.4ex plus 0.2ex minus 0.1ex}{0.6ex}
\titlespacing{\subsection}{0pt}{0.4ex plus 0.1ex minus 0.1ex}{0.5ex}
\titlespacing*{\paragraph}{0pt}{0.3ex plus 0.1ex minus 0.1ex}{0.5em}
\titleformat{\paragraph}[runin]{\normalfont\normalsize\bfseries}{\theparagraph}{0.2em}{}

\titlespacing*{\paragraph}{0pt}{0.1ex plus 0.1ex minus 0.1ex}{0.5em}
\titleformat{\paragraph}[runin]{\normalfont\normalsize\bfseries}{\theparagraph}{0.2em}{}

\begin{document}
\maketitle
\begin{abstract}
Despite the rapid progress of Vision-Language Models (VLMs), the field lacks benchmarks that rigorously diagnose their true reasoning abilities and chart meaningful progress toward human-like multimodal intelligence. Most existing evaluations focus on piecemeal or disconnected tasks, obscuring critical cognitive weaknesses and providing little insight for targeted improvement.
To address this gap, we introduce \textbf{BloomBench}, part of the \textit{Almieyar} benchmarking series, the first cognitively human-grounded, bilingual (English–Arabic) multimodal benchmark for VLMs. Grounded in Bloom’s Taxonomy, BloomBench systematically evaluates six levels of cognition (Remember, Understand, Apply, Analyze, Evaluate, Create) through carefully designed image–question–answer tasks. Built with a semi-automated pipeline and validated through a stratified hybrid quality assurance protocol, it ensures scalability, cultural inclusivity, and linguistic fidelity. 
Leveraging this framework, we conduct a comprehensive study of state-of-the-art VLMs to diagnose their cognitive profiles. 
Our analysis reveals a sharp cognitive asymmetry: while state-of-the-art models achieve strong performance ceilings in semantic understanding, they struggle substantially with factual recall and creative synthesis. This demonstrates that current general multimodal proficiency masks deeper limitations in specific cognitive layers.
Furthermore, our study highlights a critical performance gap between Arabic and English, exposing limitations in current cross-lingual multimodal reasoning. These findings establish a foundation for developing more cognitively aligned and inclusive VLMs.
The benchmark framework and dataset is available at: \url{https://github.com/qcri/Almieyar-Oryx-BloomBench}.
\end{abstract}

% \newcommand{\defaultfootnote}{\thefootnote}
% \renewcommand{\thefootnote}{\textasteriskcentered}
% \footnotetext{These authors contributed equally.}
% \renewcommand{\thefootnote}{\defaultfootnote}

\renewcommand{\thefootnote}{\textsection}
%\footnotetext{This author is supported by the Konrad Zuse School of Excellence in Learning and Intelligent Systems (ELIZA) through the DAAD programme Konrad Zuse Schools of Excellence in Artificial Intelligence, sponsored by the Federal Ministry of Education and Research.}

\section{Introduction}
\noindent
Advances in transformer architectures \cite{10.5555/3295222.3295349}, increasing computational resources, and the availability of large-scale training corpora \cite{naveed2024comprehensiveoverviewlargelanguage} have driven rapid progress in language modeling. Foundational Large Language Models (LLMs) \cite{NEURIPS2022_b1efde53, grattafiori2024llama3herdmodels, touvron2023llama2openfoundation, qwen2025qwen25technicalreport, anil2023palm2technicalreport} now excel at tasks such as instruction following \cite{qin-etal-2024-infobench}, reasoning \cite{10.5555/3600270.3602070}, in-context learning \cite{NEURIPS2020_1457c0d6}, and multilingual translation \cite{zhu-etal-2024-multilingual}. Despite these advances, two broad limitations persist: First, the supply of high-quality and diverse text data is finite \cite{villalobos2022willwe}, which motivates research into data-efficient and extrapolative generalization methods \cite{li2024beyond}. Second, single-modality architectures are inherently limited when processing real-world information that spans modalities (e.g., text, images, and video) and requires reasoning about cross-modal relationships \cite{goodwin2018multimodality, yin2024survey}.

% \vspace{-1em}
\paragraph{Vision-Language Models.} 
The pursuit of artificial general intelligence (AGI), combined with the aforementioned limitations, has led to the development of Vision-Language Models (VLMs). By extending LLMs to process visual inputs (e.g., images and videos), VLMs gain a more comprehensive understanding of spatial relationships, objects, scenes, and abstract concepts \cite{li2025survey, bordes2024introduction, liu2023llava, geminiteam2024geminifamilyhighlycapable, 10.5555/3618408.3619222}. Early work such as CLIP \cite{radford2021learning} catalyzed rapid advances in multimodal tasks that integrate visual and textual understanding, including visual question answering \cite{song-etal-2022-clip}, image captioning \cite{dai2023instructblip}, embodied agents \cite{ma2024survey}, and document understanding \cite{luo2024layoutllm}. Notably, models like GPT-4 \cite{openai2024gpt4technicalreport} demonstrate human-level performance by jointly processing text and images, marking a significant milestone in multimodal AI. Nevertheless, important challenges remain: current VLMs still struggle with complex visual reasoning \cite{NEURIPS2022_11332b6b}, object hallucination \cite{leng2024mitigating}, fine-grained spatial understanding \cite{daxberger2025mm}, and compositional reasoning \cite{sahin2024enhancing}. These limitations underscore the need for a comprehensive evaluation that systematically assesses multimodal capabilities.

\paragraph{Multimodal Benchmarks.} To systematically measure capabilities, LLMs and VLMs are typically evaluated on benchmarks: curated collections of tasks or questions designed to assess performance in domains such as mathematics \cite{wang2024mathhay}, programming \cite{yang2025probench}, and biology \cite{justen2025llms, phan2025humanity}. Significant progress has been made in designing domain-specific and large-scale benchmarks in recent years. However, conventional benchmarks have been criticized for relying on artificial datasets that fail to capture the complexity of human-level tasks \cite{zhong-etal-2024-agieval}. As a result, high performance on one benchmark (e.g., reading comprehension) does not necessarily translate to proficiency in other cognitive skills (e.g., arithmetic), and low performance on a benchmark does not straightforwardly reveal general weaknesses. This paradigm encourages the development of models that learn narrow, "shortcut" solutions specific to a benchmark's statistical patterns rather than acquiring robust, generalizable abilities. This makes it exceptionally difficult to diagnose the underlying cognitive strengths and weaknesses of models or to identify concrete directions for improvement, hindering a thorough understanding of their true capabilities. 

\begin{figure*}[t!]
\centering
  \includegraphics[width=\textwidth]{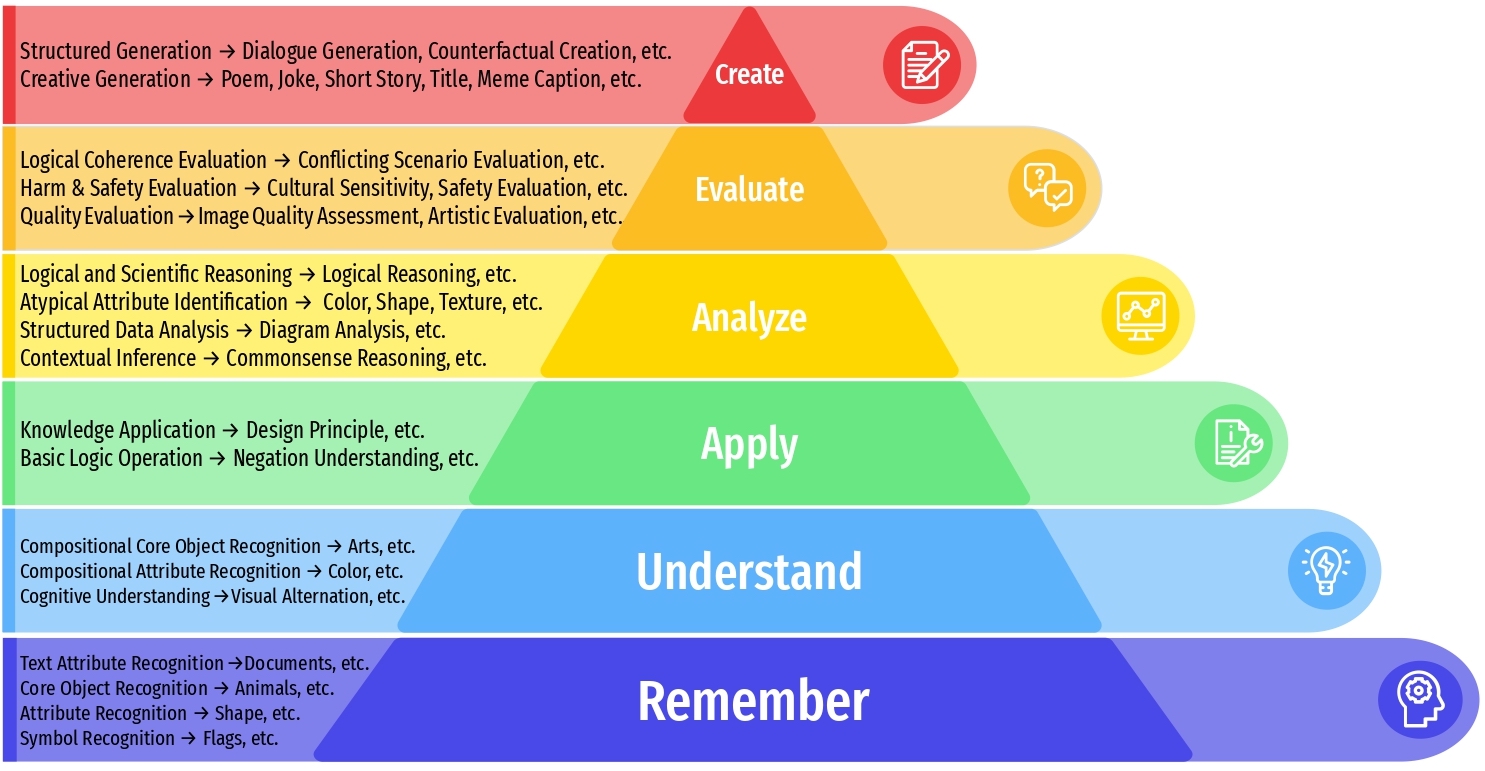}
\caption{
Hierarchical overview of the BloomBench Taxonomy. Grounded in Bloom’s cognitive framework, this hierarchy organizes multimodal tasks across six levels of cognitive complexity. Each level is further decomposed into specific task families to enable fine-grained evaluation of VLM reasoning capabilities.
\vspace{-2mm}
}
\label{fig:our_bloom}
\end{figure*}

\paragraph{Bloom's Taxonomy.}
A promising direction for more informative evaluation is to align benchmarks with cognitive science frameworks, particularly Bloom's Taxonomy \cite{zhong-etal-2024-agieval, huber2025llms}. This taxonomy, originally developed by Bloom and later revised by Anderson and Krathwohl, organizes cognitive processes into a hierarchy ranging from lower- to higher-order skills (Remember, Understand, Apply, Analyze, Evaluate, and Create) \cite{adams2015bloom, wilson2016anderson}. By framing diverse multimodal tasks within this hierarchy, assessments can move beyond surface-level accuracy to measure deeper visual reasoning abilities of VLMs. This perspective motivates our design of a benchmark that evaluates VLMs across multiple cognitive levels rather than narrow task-specific metrics. \autoref{fig:our_bloom} visualizes the taxonomy and our high-level mapping.

\paragraph{Contributions.}
In this work, \textbf{(i)} we introduce \textit{\textbf{BloomBench}}, the first multimodal, bilingual benchmark for VLMs explicitly grounded in Bloom's Taxonomy, enabling comprehensive evaluation across multiple levels of cognitive complexity. \textbf{(ii)} We address the diagnostic limitations of existing benchmarks by designing tasks that measure the depth of VLM reasoning abilities, rather than just its performance on a set of disconnected tasks. 
\textbf{(iii)} We develop a scalable generation pipeline backed by a systematic LLM-as-a-judge and human validation study on a representative subset to ensure high data quality. \textbf{(iv)} We benchmark a diverse set of state-of-the-art VLMs, providing a taxonomy-driven analysis of their strengths, weaknesses, and open challenges, and incorporate both answer-based and likelihood-based evaluation methods to reveal discrepancies between model accuracy and confidence, exposing hidden reasoning gaps.
\textbf{(v)} By incorporating a bilingual English-Arabic evaluation, our work challenges the anglocentric focus of current VLM benchmarks and enables a more inclusive assessment of how cognitive abilities generalize across diverse linguistic and cultural contexts.
Together, these contributions establish a cognitively informed framework for assessing VLM progress and guiding future research.

\section{Related Works} 
\paragraph{LLM Benchmarks.}
With the rapid growth of LLM research, benchmarks have become essential tools for tracking and comparing model capabilities. Widely adopted evaluations such as MMLU \cite{hendryckstest2021, hendrycks2021ethics}, BIG-Bench Hard \cite{suzgun2022challenging}, GSM8K \cite{cobbe2021gsm8k}, HumanEval \cite{chen2021codex}, and GPQA \cite{rein2024gpqa} provide headline scores across heterogeneous tasks. Due to their scalability and objectivity, many of these benchmarks rely on multiple-choice or short-answer formats \cite{dua-etal-2019-drop, rajpurkar2016squad, wang-etal-2018-glue, sarlin2020superglue, yang-etal-2018-hotpotqa}, which facilitate automated evaluation but limit diagnostic depth.

\paragraph{VLM Benchmarks.}
VLMs are increasingly applied in domains ranging from generative AI systems \cite{abootorabi2025generative} and retrieval-augmented generation (RAG) \cite{abootorabi-etal-2025-ask}, to education \cite{baral-etal-2025-drawedumath} and healthcare \cite{hartsock2024vision}. Despite progress, current models still face challenges in visual arithmetic \cite{huang2025why-vision}, geometric problem-solving \cite{gao2023g-llava}, and spatial reasoning tasks such as orientation, relations, and navigation \cite{stogiannidis2025mindgap, chen2024spatialvlm}. To evaluate performance in these areas, various benchmarks have been introduced.

\noindent
Early evaluation efforts were largely task-specific, focusing on specific tasks such as visual question answering (VQA) \cite{schwenk2022okvqa}, image captioning \cite{liu-etal-2021-visual}, and hallucination detection \cite{li-etal-2023-evaluating}. More recent work has sought broader coverage and higher complexity. For example, VLM2-Bench \cite{zhang-etal-2025-vlm2} evaluates fine-grained cue association across nine subtasks and 3,000 test cases, while revealing persistent weaknesses in visual grounding. MMMU \cite{yue2024mmmu} combines multiple-choice and open-ended formats to assess perception, knowledge, and reasoning. MMT-Bench \cite{ying2024mmt} spans 32 expert-level tasks requiring reasoning and localization. Other efforts focus on spatial reasoning specifically \cite{stogiannidis2025mindgap, chen2024spatialvlm}. Together, these benchmarks highlight important advances but still leave gaps in systematically evaluating higher-order reasoning.

\paragraph{Arabic-English VLM Benchmarks.}
While the VLM landscape has been predominantly English-centric, a growing body of work is developing resources for other languages, particularly Arabic. 
Initial efforts focused on translating existing English corpora, such as Violet for image captioning \cite{mohamed-etal-2023-violet} and AraCLIP for Arabic image-text alignment \cite{al2024araclip}. The availability of Arabic data was further expanded by large multilingual datasets such as WIT \cite{srinivasan2021wit} and PALI \cite{ahmadi2023pali}. 
More recently, research has advanced towards creating culturally and dialectally aware benchmarks. For instance, the Peacock model family was introduced with Henna, a benchmark assessing understanding of Arabic culture \cite{alwajih-etal-2024-peacock}, and CAMEL-Bench \cite{ghaboura2024camelbench} provides a comprehensive suite for domains ranging from handwritten document understanding to medical imaging. These efforts highlight the importance of inclusive, culturally grounded evaluation resources, particularly as dedicated Arabic-centric platforms such as Fanar continue to emerge \cite{team2025fanar, abbas2026fanar}. Our benchmark contributes to this line of work by offering bilingual evaluation (Arabic and English), with quality rigorously validated through an LLM-as-a-judge framework \cite{zheng2023judging} using Gemini 3 Pro and human to ensure high linguistic fidelity and cognitive alignment.

% \paragraph{Human Cognition-based Benchmarks.}
% To address the diagnostic limitations of task-based evaluations, researchers increasingly turn to cognitive science frameworks. In text-only settings, Bloom's Taxonomy has been used to analyze LLMs’ capabilities: \citet{huber2025llms} map popular benchmarks to the taxonomy’s six levels, finding that evaluations concentrate on mid-level skills (Apply, Analyze) while under-representing foundational (Remember) and higher-order (Evaluate, Create) skills; they also observe stronger performance of current LLMs on lower-order tasks. However, their analysis is restricted to text-only benchmarks and does not provide comprehensive coverage across all cognitive levels.
% Educators also employ Bloom's Taxonomy to design assessments that span the full skill spectrum \cite{elkins2024teachers}. 

% \vspace{-0.2em}
\paragraph{Human Cognition-based Benchmarks.}
To address the diagnostic limitations of task-based evaluations, researchers increasingly turn to cognitive science frameworks. In text-only settings, Bloom's Taxonomy helps analyze LLMs’ capabilities: \citet{huber2025llms} map popular benchmarks to the taxonomy’s six levels, finding that evaluations concentrate on mid-level skills (Apply, Analyze) while under-representing foundational (Remember) and higher-order (Evaluate, Create) skills. Beyond general analysis, recent works have operationalized the taxonomy for domain-specific evaluation. BloomAPR \cite{ma2025bloomapr} is a dynamic framework for automated program repair that transforms static benchmarks into hierarchical tasks; their findings reveal that while models effectively memorize fixes (Remember), they struggle significantly with higher-order analysis and transfer in real-world coding contexts. Similarly, BLOOMQA \cite{chen2026automated} proposed a framework for generating benchmarks from expert guidelines in practice-based domains (e.g., teaching, dietetics), observing that LLMs occasionally exhibit non-intuitive behavior by outperforming on higher-order reasoning (Analyze) while failing foundational recall tasks. Educators also employ the taxonomy to design assessments that span the full skill spectrum \cite{elkins2024teachers}. However, these efforts remain restricted to text and code modalities, lacking a comprehensive framework to evaluate cognitive depth in vision-language processing.

\noindent
Beyond Bloom's Taxonomy, ToMBench \cite{chen2024tombench} evaluates LLMs on theory-of-mind reasoning tasks, highlighting important aspects of social cognition but remaining limited to text-only settings. 
More recently, \citet{weng2025caption} proposed a multimodal framework grounded in psychological faculties such as perception, attention, and memory. While valuable, their approach lacks a cumulative hierarchy, making it difficult to assess reasoning depth in a systematic manner. These limitations highlight that their benchmark and findings are narrow in domain (synthetic objects, controlled settings) and lack linguistic and cultural diversity. In contrast, our work is explicitly grounded in Bloom’s Taxonomy, which provides a hierarchical structure for evaluating VLMs and enables a fine-grained diagnosis of cognitive abilities across the full spectrum of complexity.
% More recently, \citet{weng2025caption} proposed a multimodal framework grounded in psychological faculties such as perception, attention, and memory. While valuable, their approach lacks a cumulative hierarchy, making it difficult to assess reasoning depth in a systematic manner. In contrast, our work is explicitly grounded in Bloom’s Taxonomy, which provides a hierarchical structure for evaluating VLMs and enables a fine-grained diagnosis of cognitive abilities across the full spectrum of complexity.

% \begin{figure}[t]
%   \includegraphics[width=\columnwidth]{images/bloom_taxonomy.png}
%   \caption{Bloom's Taxonomy pyramid. The complexity of different cognition levels rises from down to up.}
%   \label{fig:bloom}
% \end{figure}

\begin{figure*}[t!]
\centering
  \includegraphics[width=\textwidth]{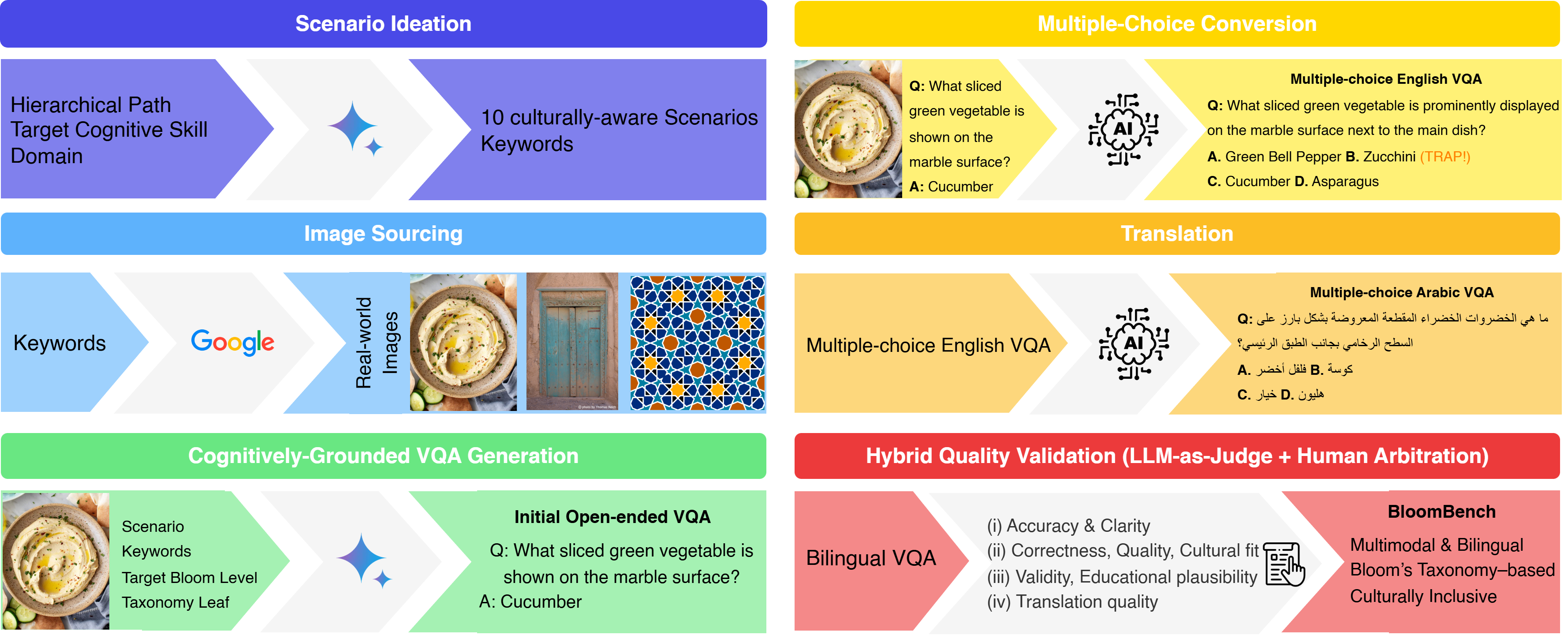}
\caption{
Overview of the BloomBench data generation pipeline. The process combines scenario ideation, cognitively-grounded VQA generation, multiple-choice conversion, translation, and hybrid quality validation of the representative subset to ensure high-quality, culturally relevant benchmark items across all Bloom’s levels.
\vspace{-2mm}
}
\label{fig:pipeline}
\end{figure*}

\section{BloomBench Methodology}
% \noindent
% This section presents BloomBench’s methodology, including its design principles, task definitions across Bloom’s Taxonomy, and data generation pipeline.

\subsection{Design Principles}
\noindent
Our goal is to design a benchmark capable of evaluating different hierarchical levels of cognitive reasoning in VLMs. Translating an abstract educational theory, such as Bloom’s Taxonomy, into a concrete, measurable, and rigorous multimodal evaluation framework requires a principled design process. 
After a careful review of prior works and an in-depth analysis of Bloom’s revised taxonomy, we define a hierarchical taxonomy of multimodal tasks, where each cognitive level (e.g., Remember, Understand, Apply, Analyze, Evaluate, Create) is further decomposed into sub-levels and specific task types. This structure allows for a comprehensive and fine-grained assessment of VLM capabilities. The design of BloomBench is guided by the following principles:

\paragraph{Cognitive Completeness.}
We aim for a holistic evaluation by providing comprehensive coverage across the entire spectrum of Bloom's Taxonomy. Unlike benchmarks that focus on a narrow range of skills, BloomBench is explicitly designed to assess the full depth of a model’s multimodal reasoning, from foundational abilities like remembering and understanding to the highest levels of evaluating and creating. This principle ensures not only comprehensive coverage but also interpretability in understanding how different reasoning abilities contribute to overall multimodal cognition.
% The taxonomy is designed to cover the entire spectrum of cognitive complexity, enabling a holistic evaluation of a model’s multimodal reasoning depth—from basic recognition to high-level abstraction and creativity.

% \subsubsection{Hierarchical Dependence}
\paragraph{Hierarchical Dependence.}
Bloom’s Taxonomy is a cumulative hierarchy in which higher-order reasoning builds on more basic skills. We adopt this structure as an organizing scaffold for BloomBench: advanced tasks are designed to presuppose competence in foundational ones. While this mirrors human cognition, we do not assume VLMs will follow the same progression. Instead, the hierarchy offers a principled way to assess where models align with or diverge from expected cognitive trajectories.
% Our tasks are designed to reflect the cumulative structure of Bloom’s Taxonomy, where competence at higher cognitive levels presupposes mastery of lower ones. This hierarchical scaffolding is crucial for meaningful evaluation. For example, a model cannot be expected to analyze the relationship between objects in an image (Analysis) without first being able to correctly identify and describe them (Remembering and Understanding). This principle allows us to probe the depth of a model's reasoning in a systematic way.

% Tasks are constructed to reflect the cumulative hierarchy of Bloom’s Taxonomy. Success at higher-order levels implicitly requires competencies from lower levels. For example, a model cannot analyze the relationship between objects in an image without first understanding their individual attributes and spatial configuration.

% \subsubsection{VLM-Specificity}
\paragraph{VLM-Specificity.}
We target core multimodal abilities over text-only reasoning with incidental visual context. Tasks require models to ground language in visual evidence, reason about spatial relations and orientation, understand visual compositionality of objects and attributes, and link abstract concepts to perceptual cues (e.g., numeracy, causality). This design isolates vision–language competence and reduces reliance on textual shortcuts.

% \subsubsection{Real-World Relevance and Scenario Diversity}
\paragraph{Real-World Relevance and Scenario Diversity.}
A key principle is to ground our evaluation in realistic, context-rich scenarios that reflect genuine human perceptual and reasoning challenges. To achieve this, our tasks draw from diverse, authentic web-sourced images covering everyday, abstract, and domain-specific contexts. Such diversity ensures that models are tested on the variability and ambiguity present in natural visual environments, conditions under which human cognition operates. By moving beyond synthetic or overly controlled settings, we assess a model’s ability to generalize its cognitive skills to complex, real-world visual understanding.

% \subsubsection{Scalability and Transparency}
\paragraph{Scalability and Transparency.}
Another core design principle of BloomBench is to ensure a scalable yet methodologically rigorous construction process. To this end, we adopt a semi-automated pipeline supported by a hybrid validation layer. This approach allows us to leverage the efficiency of automated methods for initial task generation while ensuring reliability through statistically grounded quality assurance. By employing state-of-the-art reasoning models to validate a stratified representative subset, audited by humans, we establish high-confidence quality baselines without the bottleneck of exhaustive manual verification. The result is a benchmark that is not only large-scale but also transparent, statistically validated, and reproducible by design.

% \begin{figure}[t]
%   \includegraphics[width=\columnwidth]{images/our-taxonomy.jpg}
%   \caption{Our proposed taxonomy, grounded in Bloom’s framework, including two sublevels under each Bloom level, forming a hierarchical pyramid.}
%   \label{fig:our_bloom}
% \end{figure}

\subsection{Task Design for Each Cognitive Level}
\noindent
BloomBench operationalizes the cognitive hierarchy through six task families, each corresponding to a level in Bloom's Taxonomy. These families are designed to be VLM-specific, testing the progressive development of multimodal reasoning from perception to creative synthesis. We outline the objectives and representative task types for each level. A summary of the taxonomy is shown in \autoref{fig:our_bloom}, with the complete task list in Appendix ~(\S \ref{sec:fullbloom}).
%A complete taxonomy and task list for BloomBench are provided in Appendix~\ref{sec:app_taxonomy}.

\paragraph{Remember.} This foundational level assesses perceptual recognition and factual recall. Tasks are grouped into several key areas: \textit{Core Object Recognition} (e.g., animals, scenes), \textit{Attribute Recognition} (e.g., color, texture), \textit{Activity Recognition} (e.g., interactions, professions), and the identification of symbolic and textual information via \textit{Symbol Recognition} (e.g., logos, traffic signs) and \textit{Text Attribute Recognition}.

\paragraph{Understand.} Moving beyond recognition, this level probes comprehension of relationships and compositional meaning. The main subcategories are \textit{Compositional Recognition}, which tests the understanding of multiple objects and their attributes within a scene, and \textit{Cognitive Understanding}, which requires interpreting more abstract concepts like emotions, semantic knowledge, and visual paraphrasing.

\paragraph{Apply.} This level tests the ability to use learned knowledge in novel visual contexts. Tasks fall into two primary groups: \textit{Knowledge Application}, which involves applying external concepts, such as mathematical formulas or scientific principles, to a visual input; and \textit{Basic Logic Operations}, which test the understanding of negation, word order, and coordination in a multimodal context.

\paragraph{Analyze.} Analytical tasks require a model to deconstruct a scene to infer relationships and patterns. This family is broken down into four key reasoning types: \textit{Logical and Scientific Reasoning}, \textit{Contextual Inference} (e.g., resolving ambiguity or pronouns), \textit{Structured Data Analysis} (e.g., interpreting charts and tables), and \textit{Atypical Attribute Identification} (e.g., spotting unusual colors or shapes).

\paragraph{Evaluate.} This level measures a model's capacity for judgment and critical assessment. Tasks require making and justifying evaluations, organized into three core areas: \textit{Logical Coherence Evaluation} (e.g., detecting hallucinations), \textit{Harm \& Safety Evaluation} (e.g., identifying cultural insensitivity or toxicity), and \textit{Quality Evaluation} (e.g., assessing the artistic or technical quality of an image).

\paragraph{Create.} 
At the highest level, tasks assess the ability to synthesize information to generate novel content. We operationalize this in an MCQ format as \textit{discriminative creativity}: requiring the model to evaluate potential syntheses (e.g., poem endings) and identify the one best satisfying complex constraints (rhyme, narrative coherence) over flawed distractors. This measures latent creative judgment, a critical prerequisite for generative ability, across \textit{Creative Generation} (e.g., storytelling) and \textit{Structured Creation} (e.g., designing experiments).
% At the highest cognitive level, tasks assess the ability to synthesize information to generate novel content. These tasks are divided into two distinct modes: \textit{Creative Generation}, which involves open-ended outputs like writing a poem or a story based on an image, and \textit{Structured Creation}, which requires generating content that follows specific constraints, such as designing an experiment or producing a dialogue.

\subsection{Data Generation Pipeline}
\noindent
To construct BloomBench, we designed a scalable, semi-automated data generation pipeline that integrates LLMs with rigorous human oversight. This multi-stage process, illustrated in \autoref{fig:pipeline}, ensures that each benchmark item is cognitively grounded, visually relevant, and bilingually validated. The pipeline leverages prompt engineering techniques \cite{sahoo2024systematic} and an agentic design framework \cite{plaat2025agentic} to maintain both efficiency and quality throughout the generation process. Prompts used in the data generation pipeline are provided in the Appendix~(\S \ref{sec:prompts}).

\paragraph{Scenario Ideation and Image Sourcing.}
The pipeline begins by generating a diverse set of visual scenarios for each leaf node in the BloomBench taxonomy. A high-capacity language model (Gemini 2.5 Pro \citealp{comanici2025gemini25pushingfrontier}) is prompted with the complete hierarchical path, target cognitive skill, and domain. For each skill, the model produces ten culturally aware scenarios, including Western, MENA, and Arabic contexts, alongside carefully selected keywords that result in visually concrete images and minimize textual elements. These keywords are then used to source authentic, real-world images from the web, ensuring diversity and contextual relevance throughout the dataset.

\paragraph{Cognitively-Grounded VQA Generation.}
For each sourced image, an initial open-ended visual question–answer (VQA) pair is generated. We prompt the same LLM used in the previous step with the image, the full scenario context and keywords, and a detailed description of the target Bloom's level and taxonomy leaf. The prompt is carefully crafted to ensure the resulting question is answerable solely through the visual content, reinforcing alignment with the intended cognitive skill and minimizing reliance on external knowledge or textual cues.

\paragraph{Multiple-Choice Conversion and Translation.}
Each open-ended VQA pair is then converted into a high-quality multiple-choice question (MCQ) with four answer options. A separately instruction-tuned model is prompted to act as a professional test creator, generating three plausible distractors in addition to the correct answer, with one deliberately crafted as a ``trap'' distractor to probe for deeper understanding. The complete MCQ is then translated into Modern Standard Arabic using the same LLM, with special care to maintain the item’s semantic and cognitive integrity across both languages.

% \paragraph{Quality Validation.}
% To ensure benchmark robustness, we conducted a statistically grounded validation on a representative subset of 969 samples ($\approx$1/8 of the dataset). This sampling was stratified to include at least four random examples from every one of the 106 taxonomy leaf nodes. We utilized Gemini 3 Pro, leveraging its state-of-the-art visual reasoning capabilities, to audit sample quality. The model identified only 15 samples as potentially incorrect. Subsequent human verification confirmed these flagged cases were indeed errors. This establishes a 97.42\% quality rate, providing strong statistical confidence in the dataset’s fidelity.
\paragraph{Quality Validation.}
To ensure benchmark robustness, we implemented a \textit{Hybrid Quality Validation} protocol. First, we employed an LLM-as-a-judge phase to filter out invalid or nonsensical samples. Following this, we conducted a statistically grounded validation on a representative subset of 969 samples ($\approx$1/8 of the dataset). This sampling was stratified to include at least four random examples from every one of the 106 taxonomy leaf nodes. We utilized Gemini 3 Pro, leveraging its state-of-the-art visual reasoning capabilities, to audit sample quality. The model identified only 15 samples as potentially incorrect. Subsequent human verification of the selected representative subset confirmed that only these flagged cases were indeed errors (see Appendix~\S\ref{sec:annotation_guidelines} for annotation protocol). This establishes a 98.45\% quality rate, providing strong statistical confidence in the dataset’s fidelity.

\subsubsection{Dataset Statistics} 
% todo
BloomBench contains 7,747 bilingual (English–Arabic) image–question–answer pairs across 106 distinct task types, spanning all six levels of Bloom’s Taxonomy and providing hierarchically comprehensive coverage from basic perceptual recall to high-level creative reasoning. Specifically, the dataset includes 2,948 samples for \textit{Remember}, 1,592 for \textit{Understand}, 499 for \textit{Apply}, 1,431 for \textit{Analyze}, 592 for \textit{Evaluate}, and 685 for \textit{Create}. Overall, BloomBench offers a cognitively structured benchmark for evaluating VLM reasoning across multiple levels of complexity. Full statistics are provided in Appendix~(\S\ref{sec:statistics}), and representative examples are shown in Appendix~(\S\ref{sec:examples}).

\begin{figure}[t]
\centering
  \includegraphics[width=\columnwidth]{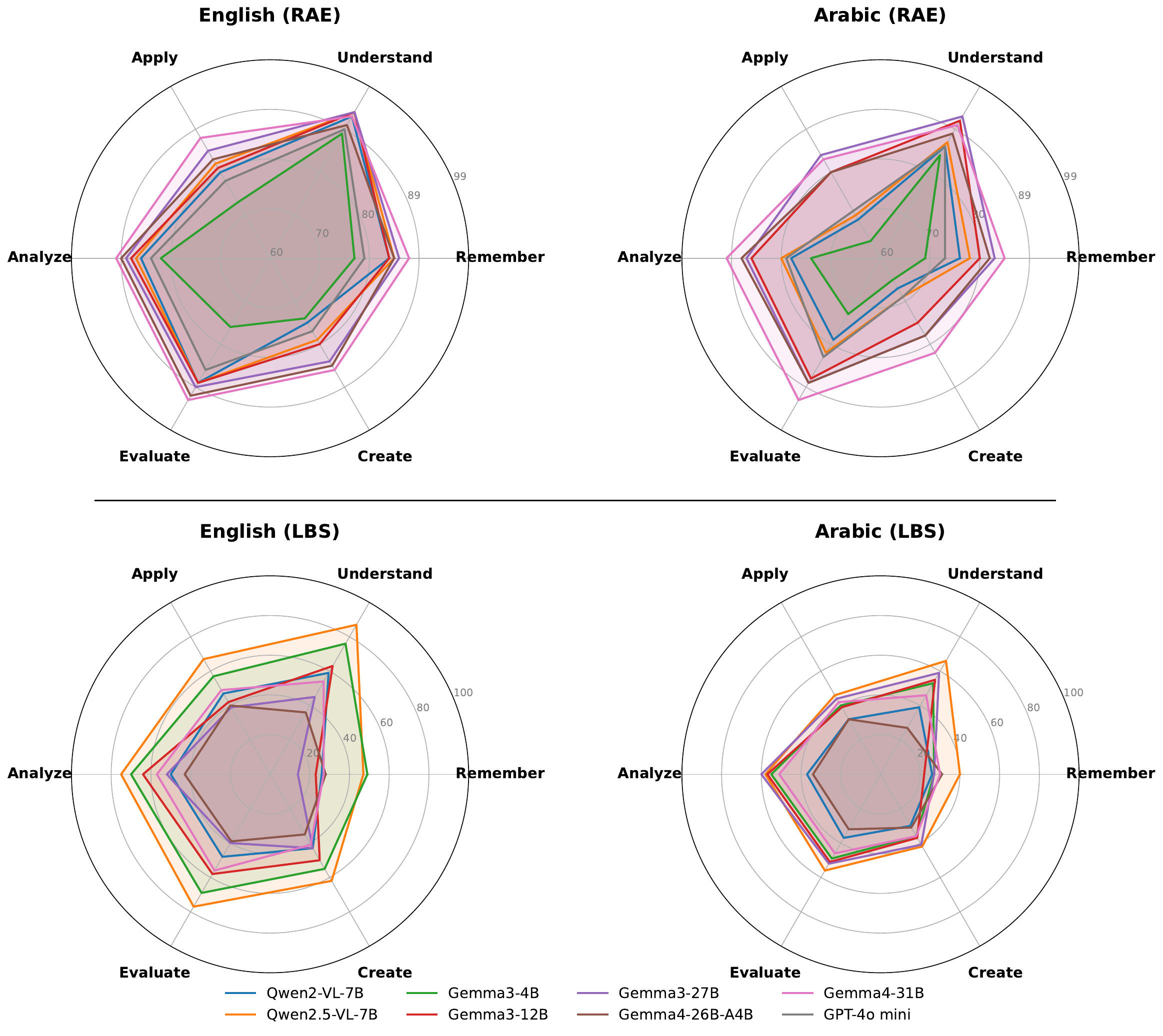}
\caption{\label{fig:vlm_bench}
Performance of different Vision-Language Models on the six BloomBench cognitive levels. The charts compare results for English (left column) and Arabic (right column) using two evaluation methods: Regex-based Answer Extraction (RAE) (top row) and Likelihood-based Scoring (LBS) (bottom row).
\vspace{-5mm}
}
\end{figure}

\section{Evaluation}
\noindent
We evaluate the performance of several representative state-of-the-art open- and closed-source VLMs on BloomBench and analyze their capabilities across different tasks and cognitive levels. We describe the evaluation setup and present several quantitative results.

\paragraph{Setup}
We evaluate a selection of representative open-source models, including Gemma 3 (4B, 12B, and 27B) \cite{gemmateam2025gemma3technicalreport}, Gemma 4 (26B-A4B and 31B) \cite{gemma4}, Qwen2.5-VL-7B \cite{qwen2025qwen25technicalreport}, and Qwen2-VL-7B \cite{wang2024qwen2vlenhancingvisionlanguagemodels}. We also include the closed-source model GPT-4o mini \cite{openai2024gpt4technicalreport} to benchmark against state-of-the-art proprietary systems. By including multiple sizes for Gemma 3 \cite{gemmateam2025gemma3technicalreport}, we aim to analyze the impact of model scale on cognitive reasoning abilities. All models are prompted with zero-shot instructions. For inference, we set the decoding temperature to 0. Following prior work in robust VLM evaluation \cite{ghahroodi2025meena, ghahroodi2024khayyam}, we employ two distinct answer extraction techniques: \textit{Regex-based Answer Extraction} and \textit{Likelihood-based Scoring}. This dual approach allows us to assess both the model's explicitly generated output and its underlying confidence distribution over the answer choices.

\paragraph{Regex-based Answer Extraction (RAE).}
This method (abbreviated as \textit{RAE}) parses the model's free-form text output to identify the selected answer choice (e.g., "A", "B", "C", or "D"). This approach simulates real-world usage where a user interprets the generated response directly. In cases where the model fails to produce a valid format, we assign a wrong choice to establish a baseline and account for catastrophic failures in instruction following.

\paragraph{Likelihood-based Scoring (LBS).} 
This method (abbreviated as \textit{LBS}) offers a more principled evaluation by directly querying the model's internal probability distribution over the possible answers, removing the dependency on output formatting. For each multiple-choice question, we compute the conditional log-probability of each answer choice given the image $I$ and the question $Q$. For each choice $C_i \in \{C_A, C_B, C_C, C_D\}$, we compute a score based on the conditional probability of the choice's token sequence $(w_1, ..., w_k)$ given the context:

{ \small
\begin{equation}
  \text{Score}(C_i) = \sum_{j=1}^{k} \log P(w_j | I, Q, w_1, ..., w_{j-1}) 
\end{equation}
}

\noindent
To ensure a fair comparison between choices of different lengths, we normalize this score by the number of tokens in the choice:
\begin{equation}
\label{eq2}
    \text{NormalizedScore}(C_i) = \frac{1}{k} \text{Score}(C_i)
\end{equation}

\noindent
The model's prediction is then the choice with the highest normalized score. This approach offers a more robust evaluation of the model's internal knowledge, independent of its ability to format the final answer:
\begin{equation}
 \text{Answer} = \underset{i \in \{A,B,C,D\}}{\arg\max} \text{NormalizedScore}(C_i)
 \end{equation}

\noindent
This likelihood-based approach offers a more direct measure of the model’s confidence in its answers, sidestepping errors from parsing, formatting, or exploiting alternative choices, and exposing reasoning signals that most conventional benchmarks overlook. 

\paragraph{Evaluation Metrics.}
For both methods, we report accuracy as the primary metric, calculated as the fraction of correctly answered questions:

{
\small
\begin{equation}                                                                                           \text{Accuracy} = \frac{1}{N} \sum_{i=1}^{N} \mathbb{I}(\text{prediction}_i = \text{ground\_truth}_i)
\end{equation}
}

\noindent
where $N$ represents the total number of test samples, $\mathbb{I}(\cdot)$ denotes the indicator function, and $\text{prediction}_i$ and $\text{ground\_truth}_i$ correspond to the model's prediction and ground truth for the $i$-th sample, respectively.

\begin{table}[t]
\centering
\small
\setlength{\tabcolsep}{3pt}
\resizebox{\columnwidth}{!}{
\begin{tabular}{l|cccc|cccc}
\toprule
\textbf{Model} & \multicolumn{4}{c|}{\textbf{English (Accuracy $\uparrow$)}} & \multicolumn{4}{c}{\textbf{Arabic (Accuracy $\uparrow$)}} \\
& \multicolumn{2}{c}{\textbf{RAE}} & \multicolumn{2}{c|}{\textbf{LBS}} & \multicolumn{2}{c}{\textbf{RAE}} & \multicolumn{2}{c}{\textbf{LBS}} \\
& \textit{Micro} & \textit{Macro} & \textit{Micro} & \textit{Macro} & \textit{Micro} & \textit{Macro} & \textit{Micro} & \textit{Macro} \\
\midrule
Qwen2-VL-7B     & 0.854 & 0.845 & 0.421 & 0.455 & 0.773 & 0.758 & 0.326 & 0.335 \\
Qwen2.5-VL-7B   & 0.869 & 0.860 & \textbf{0.654} & \textbf{0.692} & 0.792 & 0.777 & \textbf{0.503} & \textbf{0.513} \\
Gemma3-4B       & 0.796 & 0.785 & 0.609 & 0.627 & 0.729 & 0.715 & 0.408 & 0.433 \\
Gemma3-12B      & 0.867 & 0.860 & 0.450 & 0.500 & 0.836 & 0.835 & 0.398 & 0.435 \\
Gemma3-27B      & 0.883 & 0.881 & 0.336 & 0.387 & 0.859 & 0.856 & 0.440 & 0.472 \\
Gemma4-26B-A4B  & 0.876 & 0.877 & 0.347 & 0.368 & 0.846 & 0.843 & 0.309 & 0.312 \\
Gemma4-31B      & \textbf{0.898} & \textbf{0.898} & 0.430 & 0.473 & \textbf{0.876} & \textbf{0.875} & 0.397 & 0.418 \\
GPT-4o mini     & 0.824 & 0.823 & N/A   & N/A   & 0.769 & 0.768 & N/A   & N/A   \\
\bottomrule
\end{tabular} }
\caption{\textbf{Overall accuracy ($\uparrow$) of VLMs on BloomBench.} We report both \textbf{Micro} (standard) accuracy and \textbf{Macro} (balanced) accuracy to account for class imbalance across cognitive levels. N/A: Closed-source models do not support LBS calculation.}
\label{tab:results}
\vspace{-5mm}
\end{table}

\begin{table*}[t]
\centering
\scriptsize
\setlength{\tabcolsep}{2pt}
\resizebox{\textwidth}{!}{
\begin{tabular}{l|cc|cc|cc|cc|cc|cc|cc|cc|cc|cc|cc|cc}
\toprule
 & \multicolumn{12}{c|}{\textbf{English (Accuracy $\uparrow$)}} & \multicolumn{12}{c}{\textbf{Arabic (Accuracy $\uparrow$)}} \\
\textbf{Model} 
& \multicolumn{2}{c|}{Remember} & \multicolumn{2}{c|}{Understand} & \multicolumn{2}{c|}{Apply} & \multicolumn{2}{c|}{Analyze} & \multicolumn{2}{c|}{Evaluate} & \multicolumn{2}{c|}{Create} 
& \multicolumn{2}{c|}{Remember} & \multicolumn{2}{c|}{Understand} & \multicolumn{2}{c|}{Apply} & \multicolumn{2}{c|}{Analyze} & \multicolumn{2}{c|}{Evaluate} & \multicolumn{2}{c}{Create} \\
& \textbf{RAE} & \textbf{LBS} & \textbf{RAE} & \textbf{LBS} & \textbf{RAE} & \textbf{LBS} & \textbf{RAE} & \textbf{LBS} & \textbf{RAE} & \textbf{LBS} & \textbf{RAE} & \textbf{LBS}
& \textbf{RAE} & \textbf{LBS} & \textbf{RAE} & \textbf{LBS} & \textbf{RAE} & \textbf{LBS} & \textbf{RAE} & \textbf{LBS} & \textbf{RAE} & \textbf{LBS} & \textbf{RAE} & \textbf{LBS} \\
\midrule
Qwen2-VL-7B    & 0.84 & 0.26 & 0.93 & 0.59 & 0.80 & 0.47 & 0.86 & 0.50 & 0.89 & 0.48 & 0.75 & 0.43 & 0.76 & 0.26 & 0.86 & 0.39 & 0.69 & 0.32 & 0.78 & 0.37 & 0.79 & 0.37 & 0.67 & 0.30 \\
Qwen2.5-VL-7B  & 0.85 & 0.47 & 0.94 & \textbf{0.87} & 0.82 & \textbf{0.67} & 0.87 & \textbf{0.75} & 0.89 & \textbf{0.77} & 0.79 & \textbf{0.62} & 0.78 & \textbf{0.40} & 0.87 & \textbf{0.66} & 0.70 & \textbf{0.46} & 0.80 & 0.58 & 0.82 & \textbf{0.56} & 0.69 & \textbf{0.42} \\
Gemma3-4B   & 0.77 & \textbf{0.49} & 0.89 & 0.76 & 0.73 & 0.57 & 0.82 & 0.70 & 0.76 & 0.69 & 0.74 & 0.55 & 0.69 & 0.27 & 0.84 & 0.53 & 0.64 & 0.40 & 0.74 & 0.55 & 0.73 & 0.49 & 0.65 & 0.36 \\
Gemma3-12B  & 0.84 & 0.23 & 0.94 & 0.63 & 0.81 & 0.42 & 0.88 & 0.64 & 0.89 & 0.58 & 0.80 & 0.50 & 0.80 & 0.22 & 0.92 & 0.55 & 0.80 & 0.39 & 0.86 & 0.57 & 0.88 & 0.51 & 0.75 & 0.37 \\
Gemma3-27B  & 0.86 & 0.14 & \textbf{0.94} & 0.45 & 0.85 & 0.39 & 0.89 & 0.52 & 0.90 & 0.40 & 0.84 & 0.43 & 0.83 & 0.27 & \textbf{0.93} & 0.59 & \textbf{0.84} & 0.44 & 0.87 & \textbf{0.60} & 0.89 & 0.52 & 0.78 & 0.41 \\
Gemma4-26B-A4B & 0.85 & 0.28 & 0.91 & 0.36 & 0.83 & 0.40 & 0.90 & 0.43 & 0.92 & 0.39 & 0.85 & 0.35 & 0.82 & 0.31 & 0.89 & 0.27 & 0.80 & 0.32 & 0.88 & 0.34 & 0.89 & 0.32 & 0.78 & 0.31 \\
Gemma4-31B & \textbf{0.88} & 0.27 & 0.93 & 0.54 & \textbf{0.88} & 0.49 & \textbf{0.91} & 0.57 & \textbf{0.93} & 0.56 & \textbf{0.86} & 0.41 & \textbf{0.85} & 0.30 & 0.91 & 0.46 & 0.83 & 0.42 & \textbf{0.91} & 0.51 & \textbf{0.93} & 0.46 & \textbf{0.82} & 0.36 \\
% Fanar 2.0 & 0.54 & 0.33 & 0.78 & 0.67 & 0.71 & 0.56 & 0.80 & 0.65 & 0.82 & 0.60 & 0.68 & 0.48 & 0.44 & 0.31 & 0.65 & 0.47 & 0.62 & 0.37 & 0.71 & 0.48 & 0.77 & 0.45 & 0.59 & 0.35 \\
GPT-4o mini & 0.79 & N/A  & 0.90 & N/A  & 0.78 & N/A  & 0.84 & N/A  & 0.86 & N/A  & 0.77 & N/A  & 0.73 & N/A  & 0.86 & N/A  & 0.71 & N/A  & 0.79 & N/A  & 0.83 & N/A  & 0.69 & N/A  \\
\bottomrule
\end{tabular}
}
\caption{Accuracy ($\uparrow$) of VLMs across BloomBench cognitive levels. Results reported for Regex-based Answer Extraction (RAE) and Likelihood-based Scoring (LBS). Bold denotes the best results for each metric–language pair. N/A: Closed-source models do not support LBS calculation.}
\label{tab:multicol_lang_state_metricric}
\vspace{-5mm}
\end{table*}

\section{Discussion}
\label{sec:discussion}
\noindent
The overall results of all evaluated models on BloomBench using both evaluation methods are presented in \autoref{tab:results}, while \autoref{tab:multicol_lang_state_metricric} and \autoref{fig:vlm_bench} detail their performance across the six cognitive levels of our taxonomy. Task-specific results are provided in Appendix~(\S\ref{sec:details}). The key findings are summarized below.

\noindent
\paragraph{Insights and Challenges from BloomBench.}
BloomBench presents substantial challenges. While Gemma 4 31B achieves the state-of-the-art performance in Regex-based accuracy (89.8\% English / 87.6\% Arabic), overtaking Qwen2.5-VL, it notably struggles under the LBS evaluation. Performance on Arabic tasks generally lags behind English; however, the Gemma 3 family demonstrates remarkable cross-lingual consistency, with the 27B model showing a minimal performance drop between languages.

\paragraph{Coverage Comparison with Existing Benchmarks.}
To concretely illustrate BloomBench's diagnostic value relative to existing resources, we examined MMMU \cite{yue2024mmmu} as an established example of current evaluation standards. Specifically, we mapped 1,080 of its samples onto the BloomBench taxonomy using Gemini 3 Flash \cite{gemini3flash} as a judge.
The results highlight a pronounced disparity: the \textit{Analyze} level alone accounts for 66.4\% of MMMU's coverage, driven almost entirely by Math Reasoning (344 samples), Table Analysis (85), and Chart Analysis (68), while the \textit{Create} and \textit{Evaluate} levels combined represent under 1.1\% of the dataset. Forty-five taxonomy leaf nodes, such as Ambiguity Resolution, Toxicity Detection, and Dialogue Generation, have zero representation in MMMU. This confirms that while MMMU excels at evaluating 
expert domain knowledge and analytical reasoning, it cannot serve as a proxy for the full spectrum of multimodal cognition. A complete breakdown is provided in Appendix~(\S\ref{sec:mmmu_coverage}).

\noindent
\paragraph{Comparison Between RAE and LBS Evaluation.}
Previous benchmarks typically rely on RAE, but our experiments reveal that LBS exposes deeper weaknesses. While critics might attribute LBS difficulty to scoring artifacts, the disparate impact across models contradicts this.
While Qwen2.5-VL maintains relative stability ($0.869$ RAE $\rightarrow$ $0.654$ LBS), the Gemma 3 family exhibits an inverse scaling trend in LBS. Most strikingly, Gemma 3 27B achieves the highest RAE accuracy ($0.883$) yet suffers the most severe drop in LBS ($0.336$).
This divergence suggests that Gemma3 relies more on surface-level pattern recognition, while the Qwen family demonstrates superior internal consistency. Thus, the two metrics provide complementary perspectives: RAE simulates real-world usage, while LBS validates the model’s underlying reasoning confidence.

% Previous benchmarks typically rely on RAE or similar methods, but our experiments reveal that the LBS method exposes deeper weaknesses. Notably, Gemma models show a sharp performance drop under LBS, suggesting a reliance on pattern recognition and answer selection rather than intrinsic reasoning. In contrast, Qwen2.5-VL maintains strong and stable results across both methods, reflecting better internal consistency, confidence, and reduced dependency on output formatting. Together, RAE and LBS provide complementary perspectives: RAE simulates real-world usage, while LBS probes the model’s underlying reasoning confidence.

\noindent
\paragraph{Cross-Linguistic Performance.}
% As expected, performance is consistently higher in English than in Arabic across all models and metrics. Among the evaluated systems, the Gemma-3 family shows the smallest drop when shifting to Arabic, reflecting stronger cross‑lingual generalization as noted by the authors \cite{gemmateam2025gemma3technicalreport}, while the Qwen models exhibit more pronounced declines.
As expected, performance is consistently higher in English than in Arabic. We note that this gap is amplified by tokenization bias in LBS: Arabic’s higher morphological fertility leads to more tokens per word, disproportionately penalizing the length-normalized score (Equation \ref{eq2}). 
To rigorously disentangle genuine reasoning gaps from metric sensitivity, we conducted a controlled ablation using Spanish, a high-resource language with distinct tokenization properties from English, confirming that LBS scores drop significantly even when RAE remains near-parity with English, validating that the Arabic LBS gap reflects a compound effect of tokenization fertility and lower non-English probability priors (see Appendix~(\S\ref{sec:lbs_ablation})).
Despite this, the Gemma-3 family shows the smallest drop when shifting to Arabic, reflecting stronger cross-lingual generalization \cite{gemmateam2025gemma3technicalreport}, while Qwen models exhibit more pronounced declines.

\noindent
\paragraph{Model Size Effect on Performance.}
We observe a distinct decoupling of metrics as model size increases. Under RAE, performance scales predictably with size (27B > 12B > 4B). However, under LBS, we observe an inverse scaling phenomenon within the Gemma 3 family, where the largest 27B model yields the lowest likelihood scores. This suggests that increasing model scale and instruction tuning intensity may improve deterministic answer generation at the cost of the raw probabilistic calibration required for LBS.

\noindent
\paragraph{Cognitive-Level Analysis.}
Performance trends across Bloom’s hierarchy reveal non-linear patterns. Models demonstrate near-ceiling proficiency in \textit{Understand} and \textit{Evaluate} (achieving $>0.88$ RAE), indicating that discriminative visual reasoning is highly advanced in state-of-the-art VLMs. However, this competence does not transfer to generative tasks, with performance degrading significantly on \textit{Apply} and \textit{Create}. Surprisingly, \textit{Remember} also performs poorly under LBS. This discrepancy reflects current VLM training biases: models are optimized for semantic association rather than factual recall or creative synthesis. Thus, BloomBench’s diagnostic value lies in quantifying this cognitive asymmetry, revealing that high discriminative accuracy often masks deeper deficiencies in precise reasoning and generation.
% Performance trends across Bloom’s hierarchy reveal non-linear patterns. Models excel at \textit{Understand} and \textit{Evaluate} but perform worse on \textit{Apply} and \textit{Create}. Surprisingly, \textit{Remember}, typically regarded as the easiest level, performs poorly under LBS. This discrepancy reflects current VLM training biases: they are optimized for semantic association rather than factual recall. As such, they handle conceptual understanding and discriminative reasoning better than structured memorization or creative synthesis. Moreover, LBS amplifies token-level sensitivity: brief factual answers in \textit{Remember} tasks are heavily penalized by slight lexical shifts, whereas longer answers in \textit{Understand} or \textit{Evaluate} dilute these penalties.
% These findings suggest that existing VLMs demonstrate strong semantic fluency and plausibility judgment but lack robust recall and generative reasoning. BloomBench thus exposes cognitive asymmetries that conventional benchmarks overlook.

\paragraph{Cognitive-Level Cross-Linguistic Analysis.}
We observe distinct cross-lingual trends based on the RAE metric. First, the \textit{Understand} level exhibits the strongest alignment across languages, showing the lowest average degradation, indicating that core semantic comprehension is highly transferable. In contrast, across all evaluated model families, we observe a consistent and pronounced performance degradation in the \textit{Create} level when shifting from English to Arabic. This trend persists even in the Gemma 3 family, which is otherwise renowned for its multilingual capabilities, highlighting that while semantic understanding transfers effectively across languages, the high-order generative synthesis required for creation remains a significant cross-lingual bottleneck. Beyond creation, we identify a sharp divergence in procedural reasoning (\textit{Apply}). The Qwen families and GPT-4o mini suffer substantial drops in this category (e.g., $12\%$ degradation for Qwen2.5-VL). This failure in procedural reasoning is further corroborated by Likelihood-based Scoring (LBS): Qwen2.5-VL, which struggled in RAE, shows an even steeper decline in likelihood confidence ($0.67 \rightarrow 0.46$) for \textit{Apply}, confirming that its procedural failures stem from fundamental reasoning gaps rather than formatting errors. Conversely, the larger Gemma models (12B and 27B) maintain remarkable stability in this category, with negligible performance loss. However, this robustness is strictly scale-dependent in these models: while Gemma 3 12B and 27B effectively bridge the cross-lingual gap, the 4B variant suffers significant losses across all cognitive levels (e.g., $9\%$ drop in \textit{Apply} in comparison to $1\%$ drop of 12B and 27B in the same category). This underscores that robust cross-lingual alignment in complex reasoning tasks, specifically procedural application and creative synthesis, is likely an emergent property of model scale.

\section{Conclusion}
\noindent
We present BloomBench, a cognitively informed, bilingual benchmark designed to evaluate VLMs through the hierarchical lens of Bloom’s Taxonomy. By integrating cognitive theory with a semi-automated, hybrid-verified data pipeline, BloomBench enables systematic and interpretable assessment of multimodal reasoning across six levels of cognition. Our analyses reveal both the strengths and persistent weaknesses of current VLMs, underscoring the value of cognition-driven evaluation for guiding model development. Future work should expand this framework with more challenging and diverse task types, particularly at higher cognitive levels where complex reasoning extends beyond the current MQA format, and explore adaptive difficulty scaling to better capture the evolving multimodal reasoning abilities of next-generation models.

\section{Limitations}
\noindent
While this work introduces a comprehensive, cognitively grounded benchmark for evaluating Vision-Language Models, some limitations should be acknowledged. First, computational resource constraints, including limited access to large-scale GPU infrastructure and the high costs of proprietary model APIs, restricted the breadth of models we could evaluate. Consequently, several noteworthy VLMs from recent literature were not included in our experiments. We attempted to ensure diversity by selecting models spanning different architectural families and scales, but future work should expand the evaluation to encompass a broader range of state-of-the-art systems to provide more comprehensive insights into the current landscape of VLM capabilities. 
Moreover, because our pipeline relies on agent‑driven generation over rich public web data, the benchmark can naturally improve as stronger agents (e.g., Gemini 3 Pro) and more compute become available. Access to newer high‑performing models and larger GPU resources would enable future iterations of BloomBench to produce higher‑quality and more accurate multimodal items.

\noindent
Second, all questions in BloomBench are presented in multiple-choice format. While this design choice facilitates standardized evaluation and enables robust automated scoring, it may not fully capture the range of reasoning abilities required in open-ended, real-world scenarios. Incorporating additional question formats, such as fill-in-the-blank, short-answer generation, or multi-step reasoning tasks, could provide richer diagnostic information about model capabilities across cognitive levels. Future iterations of the benchmark could explore these alternative formats to complement the current evaluation framework.

\noindent
Finally, while we conducted comprehensive validation on a stratified subset ($N{\approx}1k$), we did not manually verify every item in the full dataset. However, the high agreement rate ($>97\%$) across all 106 taxonomy nodes provides strong statistical confidence in the benchmark's overall reliability.
\noindent

\section{Ethical Considerations.}
\label{sec:ethics}

\paragraph{Data Provenance and Licensing.} 
The images in BloomBench are sourced from publicly available repositories. To strictly adhere to copyright laws and fair use principles, we do not host or redistribute the image files directly. Instead, we release the dataset as a collection of image URLs accompanied by a download script. This ensures that users retrieve content from the original sources, a standard practice in recent vision-language research to respect the intellectual property and distribution rights of content creators \cite{deitke2025molmo}.

% \paragraph{Human Annotation and Fair Compensation.} 
% Our dataset construction involved human verification to ensure the quality of the cognitive mapping to Bloom's Taxonomy. We hired annotators with proficiency in both English and Arabic. We adhered to fair labor practices; annotators were compensated at a rate of \textbf{[INSERT RATE, e.g., \$15/hour]}, which exceeds the local minimum wage. All participants were informed of the data usage and provided consent prior to participation. No personally identifiable information (PII) of the annotators was collected.

\paragraph{Content Safety.} 
Given that visual data is scraped from the web, we implemented a strict filtering pipeline. We utilized both automated safety classifiers and manual inspection to remove any images containing offensive or violent content. We believe this dataset is safe for research use; however, users should exercise standard caution when utilizing web-crawled data.

\paragraph{Broader Impact.} 
BloomBench aims to advance the cognitive evaluation of LVLMs. We anticipate this will help the community move beyond surface-level pattern matching toward genuine multimodal reasoning. We do not foresee immediate negative societal impacts, but we encourage researchers to use this benchmark to identify and mitigate failures in safety-critical reasoning applications.

\section*{Acknowledgments}
\noindent
We would like to express our sincere gratitude to Mohamed Hefeeda, 
lead of the Fanar Multimodal Team at the Qatar Computing Research Institute, for his invaluable feedback and guidance throughout this work. We also extend our thanks to Hamza Aldaghstany, Mohammad Amin 
Sadeghi, and Mohamed Eltabakh, as well as the broader Fanar team, 
for their contributions and support in developing this benchmark. 
Their collective efforts were instrumental in bringing BloomBench to this stage.
The first author was also supported by the Konrad Zuse School of Excellence in Learning and Intelligent Systems (ELIZA) through the DAAD programme Konrad Zuse Schools of Excellence in Artificial Intelligence, sponsored by the Federal Ministry of Education and Research.

\bibliography{acl_latex}

\appendix
\FloatBarrier

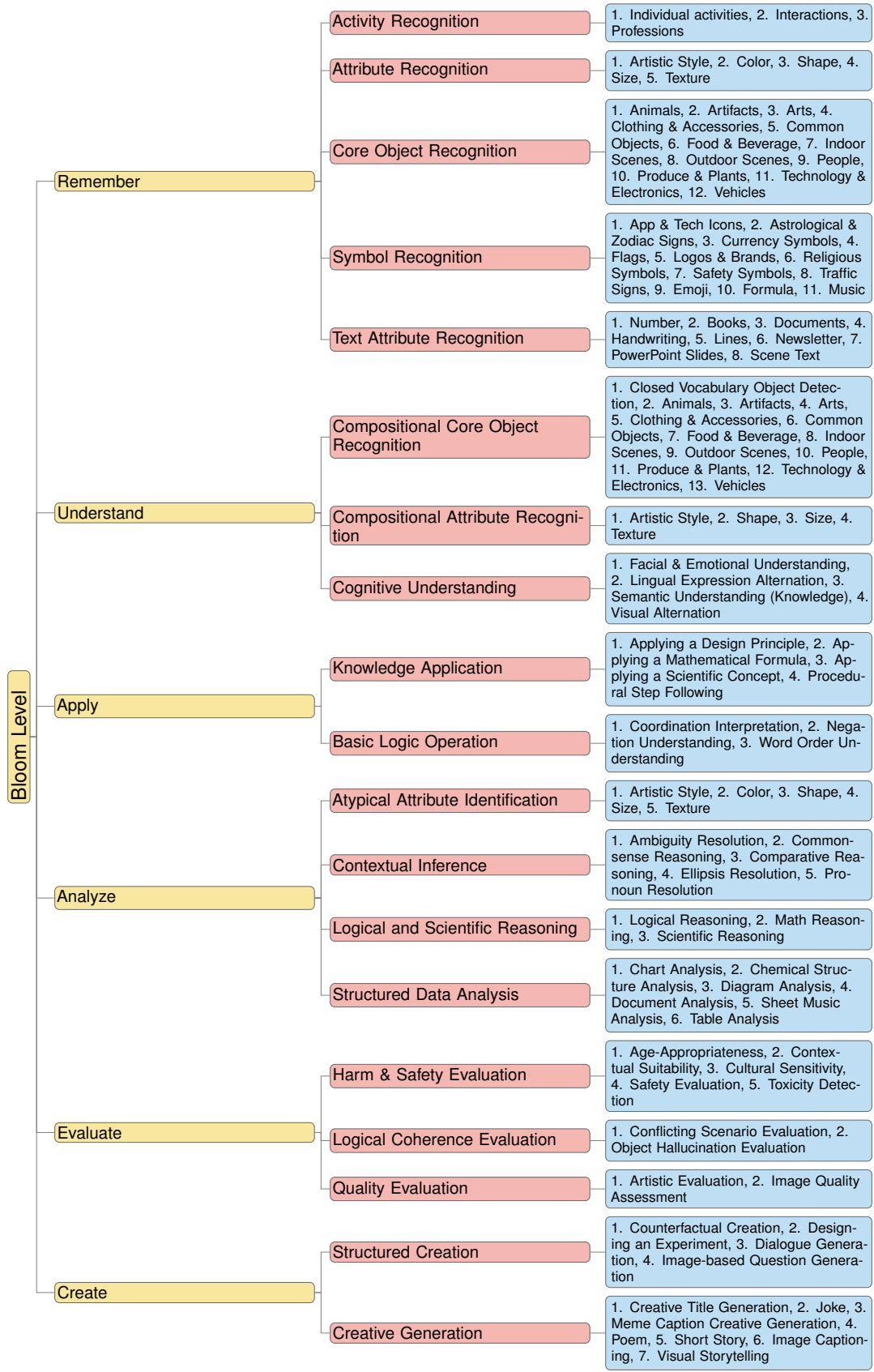
\begin{figure*}[htbp]

\centering

% ---- COLOR DEFINITIONS ----
\definecolor{catA}{HTML}{F9E79F} % Yellow (Main category)
\definecolor{catB}{HTML}{F5B7B1} % Light red (Subcategory)
\definecolor{catC}{HTML}{AED6F1} % Light blue (Leaf)
% ---- TREE ----
\section{Full BloomBench Taxonomy}
\label{sec:fullbloom}

\resizebox{!}{0.9\textheight}{%
\begin{forest}
forked edges,
for tree={
    child anchor=west,
    parent anchor=east,
    grow'=east,
    anchor=west,
    base=left,
    font=\sffamily\Large, % increase font here
    rectangle,
    rounded corners,
    draw=black!60,
    edge+={black!50, line width=0.6pt},
    s sep=7pt,
    inner xsep=0.2em,
    inner ysep=0.3em,
    text width=20em,
},
ver/.style={
    fill=catA,
    rotate=90,
    child anchor=north,
    parent anchor=south,
    anchor=center,
    text width=10em,
    font=\sffamily\LARGE
},
leaf/.style={
    fill=catC,
    inner sep=5pt,
    draw,
    fill opacity=.8,
    text opacity=1,
    text width=20em,
    font=\sffamily\large % bigger font for leaves
},
[Bloom Level, ver
    [Remember, fill=catA
        [Activity Recognition, fill=catB
            [{1. Individual activities, 2. Interactions, 3. Professions}, leaf]
        ]
        [Attribute Recognition, fill=catB
            [{1. Artistic Style, 2. Color, 3. Shape, 4. Size, 5. Texture}, leaf]
        ]
        [Core Object Recognition, fill=catB
            [{1. Animals, 2. Artifacts, 3. Arts, 4. Clothing \& Accessories, 5. Common Objects, 6. Food \& Beverage, 7. Indoor Scenes, 8. Outdoor Scenes, 9. People, 10. Produce \& Plants, 11. Technology \& Electronics, 12. Vehicles}, leaf]
        ]
        [Symbol Recognition, fill=catB
            [{1. App \& Tech Icons, 2. Astrological \& Zodiac Signs, 3. Currency Symbols, 4. Flags, 5. Logos \& Brands, 6. Religious Symbols, 7. Safety Symbols, 8. Traffic Signs, 9. Emoji, 10. Formula, 11. Music}, leaf]
        ]
        [Text Attribute Recognition, fill=catB
            [{1. Number, 2. Books, 3. Documents, 4. Handwriting, 5. Lines, 6. Newsletter, 7. PowerPoint Slides, 8. Scene Text}, leaf]
        ]
    ]
    [Understand, fill=catA
        [Compositional Core Object Recognition, fill=catB
            [{1. Closed Vocabulary Object Detection, 2. Animals, 3. Artifacts, 4. Arts, 5. Clothing \& Accessories, 6. Common Objects, 7. Food \& Beverage, 8. Indoor Scenes, 9. Outdoor Scenes, 10. People, 11. Produce \& Plants, 12. Technology \& Electronics, 13. Vehicles}, leaf]
        ]
        [Compositional Attribute Recognition, fill=catB
            [{1. Artistic Style, 2. Shape, 3. Size, 4. Texture}, leaf]
        ]
        [Cognitive Understanding, fill=catB
            [{1. Facial \& Emotional Understanding, 2. Lingual Expression Alternation, 3. Semantic Understanding (Knowledge), 4. Visual Alternation}, leaf]
        ]
    ]
    [Apply, fill=catA
        [Knowledge Application, fill=catB
            [{1. Applying a Design Principle, 2. Applying a Mathematical Formula, 3. Applying a Scientific Concept, 4. Procedural Step Following}, leaf]
        ]
        [Basic Logic Operation, fill=catB
            [{1. Coordination Interpretation, 2. Negation Understanding, 3. Word Order Understanding}, leaf]
        ]
    ]
    [Analyze, fill=catA
        [Atypical Attribute Identification, fill=catB
            [{1. Artistic Style, 2. Color, 3. Shape, 4. Size, 5. Texture}, leaf]
        ]
        [Contextual Inference, fill=catB
            [{1. Ambiguity Resolution, 2. Commonsense Reasoning, 3. Comparative Reasoning, 4. Ellipsis Resolution, 5. Pronoun Resolution}, leaf]
        ]
        [Logical and Scientific Reasoning, fill=catB
            [{1. Logical Reasoning, 2. Math Reasoning, 3. Scientific Reasoning}, leaf]
        ]
        [Structured Data Analysis, fill=catB
            [{1. Chart Analysis, 2. Chemical Structure Analysis, 3. Diagram Analysis, 4. Document Analysis, 5. Sheet Music Analysis, 6. Table Analysis}, leaf]
        ]
    ]
    [Evaluate, fill=catA
        [Harm \& Safety Evaluation, fill=catB
            [{1. Age-Appropriateness, 2. Contextual Suitability, 3. Cultural Sensitivity, 4. Safety Evaluation, 5. Toxicity Detection}, leaf]
        ]
        [Logical Coherence Evaluation, fill=catB
            [{1. Conflicting Scenario Evaluation, 2. Object Hallucination Evaluation}, leaf]
        ]
        [Quality Evaluation, fill=catB
            [{1. Artistic Evaluation, 2. Image Quality Assessment}, leaf]
        ]
    ]
    [Create, fill=catA
        [Structured Creation, fill=catB
            [{1. Counterfactual Creation, 2. Designing an Experiment, 3. Dialogue Generation, 4. Image-based Question Generation}, leaf]
        ]
        [Creative Generation, fill=catB
            [{1. Creative Title Generation, 2. Joke, 3. Meme Caption Creative Generation, 4. Poem, 5. Short Story, 6. Image Captioning, 7. Visual Storytelling}, leaf]
        ]
    ]
]
\end{forest}%
} % end resizebox

\caption{
The complete hierarchical structure of the BloomBench taxonomy. The diagram illustrates the decomposition of Bloom's six cognitive levels into specific sub-categories and fine-grained task types, defining the full scope of multimodal capabilities evaluated in the benchmark.
}
\label{fig:full-bloom-taxonomy}
\end{figure*}

\clearpage
\onecolumn % Continue in one-column mode for the appendix tables
\section{Statistics}
\label{sec:statistics}

% --- TABLE 1: Distribution ---
\begin{center}
    \captionof{table}{Distribution of Items by Bloom’s Taxonomy Category}
    \label{tab:dist_bloom}
    \begin{tabular}{l r}
    \hline
    \textbf{Bloom Category} & \textbf{Count} \\
    \hline
    Remember & 2,948 \\
    Understand & 1,592 \\
    Analyze & 1,431 \\
    Create & 685 \\
    Evaluate & 592 \\
    Apply & 499 \\
    \hline
    \textbf{Total} & \textbf{7,747} \\
    \hline
    \end{tabular}
\end{center}

\vspace{20pt}

% --- TABLE 2: Remember ---
\begin{center}
    \captionof{table}{Hierarchical Statistics — Bloom’s Level: Remember}
    \label{tab:stats_remember}
    \small
    \begin{tabular}{l l r}
    \hline
    \textbf{Subcategory} & \textbf{Sub-subcategory} & \textbf{Count} \\
    \hline
    core object recognition & clothing \& accessories & 95 \\
    attribute recognition & color & 95 \\
    core object recognition & indoor scenes & 93 \\
    symbol recognition & traffic signs & 93 \\
    core object recognition & artifacts & 92 \\
    core object recognition & people & 90 \\
    symbol recognition & flags & 89 \\
    core object recognition & outdoor scenes & 88 \\
    core object recognition & vehicles & 87 \\
    core object recognition & produce \& plants & 87 \\
    core object recognition & common objects & 85 \\
    text attribute recognition & books & 85 \\
    attribute recognition & shape & 83 \\
    text attribute recognition & lines & 83 \\
    attribute recognition & texture & 82 \\
    core object recognition & arts & 81 \\
    core object recognition & animals & 80 \\
    symbol recognition & music & 79 \\
    core object recognition & food \& beverage & 79 \\
    symbol recognition & safety symbols & 78 \\
    attribute recognition & artistic style & 77 \\
    text attribute recognition & handwriting & 75 \\
    attribute recognition & size & 75 \\
    core object recognition & technology \& electronics & 74 \\
    text attribute recognition & scene text & 73 \\
    text attribute recognition & number & 73 \\
    symbol recognition & emoji & 69 \\
    symbol recognition & religious symbols & 68 \\
    text attribute recognition & documents & 66 \\
    text attribute recognition & newletter & 63 \\
    symbol recognition & formula & 62 \\
    symbol recognition & astrological \& zodiac signs & 62 \\
    text attribute recognition & power point slides & 61 \\
    activity recognition & individual activities & 61 \\
    activity recognition & interactions & 57 \\
    activity recognition & professions & 56 \\
    symbol recognition & logos \& brands & 53 \\
    symbol recognition & currency symbols & 51 \\
    symbol recognition & app \& tech icons & 48 \\
    \hline
    \end{tabular}
\end{center}

\clearpage % Start Understand on a fresh page to keep it clean

% --- TABLE 3: Understand ---
\begin{center}
    \captionof{table}{Hierarchical Statistics — Bloom’s Level: Understand}
    \label{tab:stats_understand}
    \begin{tabular}{l l r}
    \hline
    \textbf{Subcategory} & \textbf{Sub-subcategory} & \textbf{Count} \\
    \hline
    compositional core object recognition & closed vocabulary object detection & 89 \\
    cognitive understanding & semantic understanding (knowledge) & 88 \\
    compositional core object recognition & food \& beverage & 85 \\
    compositional core object recognition & clothing \& accessories & 84 \\
    compositional attribute recognition & texture & 83 \\
    compositional core object recognition & outdoor scenes & 83 \\
    compositional core object recognition & arts & 82 \\
    compositional core object recognition & animals & 81 \\
    compositional attribute recognition & shape & 81 \\
    compositional attribute recognition & artistic style & 79 \\
    compositional core object recognition & artifacts & 76 \\
    compositional core object recognition & vehicles & 75 \\
    cognitive understanding & lingual expression alternation & 74 \\
    compositional core object recognition & people & 73 \\
    cognitive understanding & facial \& emotional understanding & 71 \\
    cognitive understanding & visual alternation & 71 \\
    compositional attribute recognition & size & 70 \\
    compositional core object recognition & produce \& plants & 67 \\
    compositional core object recognition & indoor scenes & 63 \\
    compositional core object recognition & common objects & 62 \\
    compositional core object recognition & technology \& electronics & 55 \\
    \hline
    \end{tabular}
\end{center}

\vspace{20pt}

% --- TABLE 4: Apply ---
\begin{center}
    \captionof{table}{Hierarchical Statistics — Bloom’s Level: Apply}
    \label{tab:stats_apply}
    \begin{tabular}{l l r}
    \hline
    \textbf{Subcategory} & \textbf{Sub-subcategory} & \textbf{Count} \\
    \hline
    basic logic operation & negation understanding & 84 \\
    knowledge application & applying a design principle & 77 \\
    knowledge application & procedural step following & 75 \\
    knowledge application & applying a scientific concept & 73 \\
    basic logic operation & word order understanding & 67 \\
    basic logic operation & coordination interpretation & 62 \\
    knowledge application & applying a mathematical formula & 61 \\
    \hline
    \end{tabular}
\end{center}

\clearpage

% --- TABLE 5: Analyze ---
\begin{center}
    \captionof{table}{Hierarchical Statistics — Bloom’s Level: Analyze}
    \label{tab:stats_analyze}
    \begin{tabular}{l l r}
    \hline
    \textbf{Subcategory} & \textbf{Sub-subcategory} & \textbf{Count} \\
    \hline
    structured data analysis & chart analysis & 93 \\
    structured data analysis & document analysis & 88 \\
    logical and scientific reasoning & logic reasoning & 86 \\
    logical and scientific reasoning & scientific reasoning & 86 \\
    contextual inference & comparative reasoning & 82 \\
    logical and scientific reasoning & math reasoning & 81 \\
    atypical attribute identification & artistic style & 81 \\
    contextual inference & commonsense reasooning & 78 \\
    structured data analysis & table analysis & 75 \\
    structured data analysis & chemical structure analysis & 73 \\
    atypical attribute identification & shape & 73 \\
    contextual inference & pronoun resolution & 71 \\
    contextual inference & ambiguity resolution & 69 \\
    contextual inference & ellipsis resolution & 68 \\
    atypical attribute identification & color & 68 \\
    atypical attribute identification & size & 67 \\
    structured data analysis & diagram analysis & 64 \\
    structured data analysis & sheet music analysis & 52 \\
    \hline
    \end{tabular}
\end{center}

\vspace{20pt}

% --- TABLE 6: Evaluate ---
\begin{center}
    \captionof{table}{Hierarchical Statistics — Bloom’s Level: Evaluate}
    \label{tab:stats_evaluate}
    \begin{tabular}{l l r}
    \hline
    \textbf{Subcategory} & \textbf{Sub-subcategory} & \textbf{Count} \\
    \hline
    quality evaluation & image quality assessment & 82 \\
    quality evaluation & artistic evaluation & 78 \\
    logical coherence evaluation & object hallucination evaluation & 76 \\
    harm \& safety evaluation & safety evaluation & 75 \\
    harm \& safety evaluation & contextual suitability & 70 \\
    harm \& safety evaluation & age-appropriateness & 63 \\
    logical coherence evaluation & conflicting scenario evaluation & 59 \\
    harm \& safety evaluation & toxicity detection & 55 \\
    harm \& safety evaluation & cultural sensitivity & 34 \\
    \hline
    \end{tabular}
\end{center}

\vspace{20pt}

% --- TABLE 7: Create ---
\begin{center}
    \captionof{table}{Hierarchical Statistics — Bloom’s Level: Create}
    \label{tab:stats_create}
    \begin{tabular}{l l r}
    \hline
    \textbf{Subcategory} & \textbf{Sub-subcategory} & \textbf{Count} \\
    \hline
    creative generation & creative title generation & 81 \\
    creative generation & image captioning & 77 \\
    structured creation & counterfactual creation & 74 \\
    creative generation & visual stroytelling & 64 \\
    structured creation & designing an experiment & 62 \\
    structured creation & image-based question generation & 60 \\
    structured creation & dialogue generation & 59 \\
    creative generation & meme caption creative generation & 58 \\
    creative generation & joke & 54 \\
    creative generation & poem & 53 \\
    creative generation & short story & 43 \\
    \hline
    \end{tabular}
\end{center}

\clearpage
\twocolumn % Optional: return to two columns if needed
\section{Cross-Lingual LBS Ablation Study}
\label{sec:lbs_ablation}

To rigorously validate our claim that tokenization fertility amplifies the Likelihood-based Scoring (LBS) gap observed in Arabic, we conducted a controlled ablation study using Spanish as an intermediate language. Spanish is a high-resource language that uses the Latin script and shares many morphological properties with English, but exhibits different tokenization characteristics (i.e., lower token-per-word ratios than Arabic, but higher than English for the models evaluated). We evaluated Qwen2.5-VL-7B \cite{qwen2025qwen25technicalreport} on a stratified representative subset of 908 samples, drawn with at least four examples from each of the 106 taxonomy leaf nodes.

\paragraph{Results.} As shown in Table~\ref{tab:lbs_ablation}, the model's deterministic accuracy (RAE) in Spanish remains high at 84.91\%, closely comparable to its English performance of 87.22\%, a gap of only 2.31 percentage points. However, the LBS score for Spanish suffers a substantially steeper decline (8.26 points below English), despite the near-parity in RAE. This dissociation between RAE and LBS in a language where the model demonstrably understands the content confirms that LBS is sensitive to the model's intrinsic lower probability priors for non-English text, a phenomenon that is further amplified by tokenization fertility.

\paragraph{Interpretation.} These findings support the interpretation that the Arabic LBS gap observed in the main experiments (Section~\ref{sec:discussion}) is a \textit{compound effect}: (i) a \textbf{metric artifact} stemming from higher token-per-word ratios in morphologically rich languages, which disproportionately penalizes the length-normalized score (Equation~\ref{eq2}), and (ii) \textbf{genuine reasoning gaps} arising from the model's weaker Arabic-language priors. Crucially, this also validates LBS as a measure of \textit{calibrated confidence density} rather than surface-level accuracy, distinguishing it from RAE as a complementary and more diagnostically sensitive evaluation signal.

\begin{table}[t!]
\centering
\small
\resizebox{\columnwidth}{!}{%
\begin{tabular}{lcccc}
\toprule
\textbf{Language} & \textbf{RAE\textsubscript{micro}} & \textbf{RAE\textsubscript{macro}} & \textbf{LBS\textsubscript{micro}} & \textbf{LBS\textsubscript{macro}} \\
\midrule
English & 87.22 & 85.76 & 67.07 & 67.06 \\
Spanish & 84.91 & 83.29 & 58.81 & 58.81 \\
Arabic  & 79.96 & 79.20 & 50.77 & 50.24 \\
\bottomrule
\end{tabular}%
}
\caption{RAE vs.\ LBS accuracy (\%) of Qwen2.5-VL-7B on a stratified subset of 908 samples across English, Spanish, and Arabic. The near-parity in RAE between English and Spanish, coupled with a significant LBS drop, isolates the effect of tokenization fertility on likelihood-based scoring from genuine reasoning capability.}
\label{tab:lbs_ablation}
\end{table}
\section{MMMU Taxonomy Coverage Analysis}
\label{sec:mmmu_coverage}

To quantitatively assess the difference in cognitive coverage between BloomBench and an 
established benchmark, we mapped 1,080 samples from MMMU \cite{yue2024mmmu} onto 
the BloomBench taxonomy using Gemini 3 Flash \cite{gemini3flash} as a judge model. Table~\ref{tab:mmmu_coverage} 
summarizes the distribution of these samples across the six levels of Bloom's Taxonomy. 
Note that 94 samples were classified as unmatched and are excluded from leaf-level counts.

\paragraph{Distributional Skew.}
The \textit{Analyze} level alone accounts for 66.4\% (717/1,080) of the total samples, 
reflecting MMMU's strong orientation toward exam-style academic problems in STEM domains. 
The three most frequent leaf nodes together account for nearly half (${\approx}46\%$) of the 
entire dataset:
\begin{itemize}
    \item \textbf{Math Reasoning:} 344 samples
    \item \textbf{Table Analysis:} 85 samples
    \item \textbf{Chart Analysis:} 68 samples
\end{itemize}

\paragraph{Gaps in Cognitive Coverage.}
Forty-five taxonomy leaf nodes received zero coverage, revealing structural gaps in MMMU's 
ability to test higher-order synthesis and judgment. The \textit{Create} and \textit{Evaluate} 
levels combined represent only ${\approx}1.0\%$ (11/1,080) of the dataset. Critical task types 
completely absent from the MMMU sample include:
\begin{itemize}
    \item \textbf{Contextual Inference:} Ambiguity Resolution, Commonsense Reasoning
    \item \textbf{Harm \& Safety Evaluation:} Contextual Suitability, Toxicity Detection
    \item \textbf{Creative Generation:} Dialogue Generation, Designing an Experiment
\end{itemize}
This comparison confirms that while MMMU serves as a robust benchmark for expert domain 
knowledge and analytical reasoning, it lacks the breadth to evaluate the full spectrum of 
multimodal cognition, particularly in creative, evaluative, and socially grounded tasks that 
BloomBench is designed to assess.

\begin{table}[t]
\centering
\small
\setlength{\tabcolsep}{4pt}
\begin{tabular}{lrrrr}
\toprule
\textbf{Bloom Level} & \textbf{Covered} & \textbf{Total} & \textbf{Coverage} & \textbf{Samples} \\
                     & \textbf{Leaves}  & \textbf{Nodes} & \textbf{(\%)}     &                  \\
\midrule
Remember   & 25 & 39 & 64.1\% & 77  \\
Understand & 12 & 21 & 57.1\% & 134 \\
Analyze    & 13 & 19 & 68.4\% & 717 \\
Apply      &  3 &  7 & 42.9\% & 47  \\
Evaluate   &  4 &  9 & 44.4\% & 5   \\
Create     &  4 & 11 & 36.4\% & 6   \\
\midrule
\textbf{Total} & \textbf{61} & \textbf{106} & \textbf{57.5\%} & \textbf{986\textsuperscript{$\dagger$}} \\
\bottomrule
\end{tabular}
\caption{Distribution of 1,080 MMMU samples across BloomBench cognitive levels. 
$^\dagger$94 samples were classified as unmatched and are excluded from leaf-level counts.}
\label{tab:mmmu_coverage}
\end{table}
\clearpage % Ends the current page
\onecolumn % Switches the layout to single column
\section{Examples}
\label{sec:examples}

% Optional: Reduce spacing if needed
\setlength{\parskip}{5pt}
\renewcommand{\baselinestretch}{1.0}

\noindent
\begin{center}
    % --- FIRST BOX ---
    \begin{examplebox}{1}
        \includegraphics[width=\linewidth]{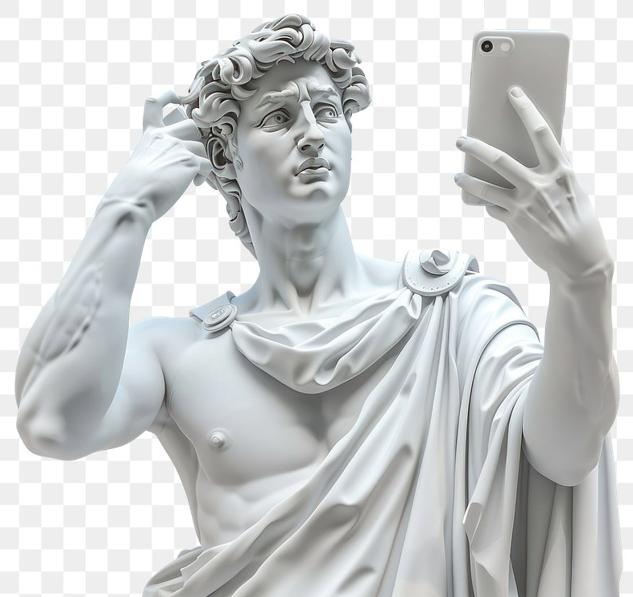}
        \vspace{4pt}
        \textbf{Levels} \\
        1: Analyze \\
        2: Atypical Attribute Identification \\
        3: Artistic Style
        \tcblower
        \textbf{Question (en):} What specific object in this image introduces a stylistic anachronism, contrasting with the classical sculpture style of the figure? 
        \vspace{4pt} \\
        \textbf{Answer:} The modern smartphone held by the figure introduces a stylistic anachronism, as it is a 21st-century device juxtaposed with a Greco-Roman or Renaissance artistic style.
        \vspace{4pt} \\
        \textbf{MCQ (en)} \\
        Which element in the image creates a striking stylistic anachronism when contrasted with the classical appearance of the sculpture?
        \begin{itemize}
            \item[(A)] The intricate details of the figure's curly hair.
            \item[(B)] The figure's intense and dramatic facial expression.
            \item[\textbf{(C)}] The modern smartphone held by the figure.
            \item[(D)] The flowing, stylized drapery worn by the figure.
        \end{itemize}
    \end{examplebox}

    \vspace{20pt} % Clear vertical gap between boxes

    % --- SECOND BOX ---
    \begin{examplebox}{1}
        \includegraphics[width=\linewidth]{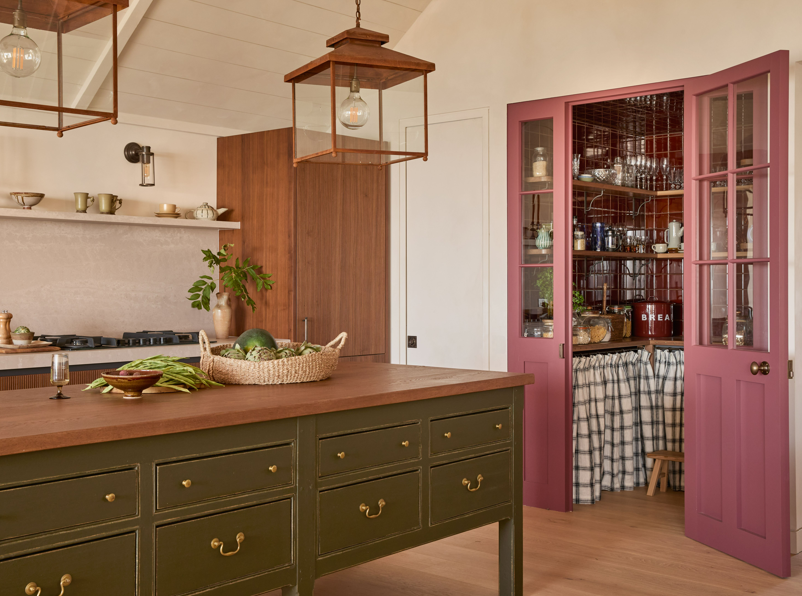}
        \vspace{4pt}
        \textbf{Levels} \\
        1: Analyze \\
        2: Contextual Inference \\
        3: Comparative Reasoning
        \tcblower
        \textbf{Question (en):} Contrast the storage approach of the pink-doored pantry with that of the tall wooden cabinets. What does this suggest about their intended contents?
        \vspace{4pt} \\
        \textbf{Answer:} The pantry uses visible storage with glass doors and open shelves, suggesting it's for frequently accessed or decorative items like glassware and dry goods. The opaque wooden cabinets offer concealed storage, likely for appliances or less aesthetic bulk supplies.
        \vspace{4pt} \\
        \textbf{MCQ (en)} \\
        Which of the following best contrasts the storage approach of the pink-doored pantry with the tall wooden cabinets, and suggests their intended contents?
        \begin{itemize}
            \item[(A)] Both storage units emphasize concealed storage; however, the pantry is for non-perishable goods, and the wooden cabinets are for cleaning supplies.
            \item[(B)] The pink-doored pantry is solely for aesthetic display, using its color to enhance the room, and the wooden cabinets are for storing large, industrial kitchen equipment.
            \item[\textbf{(C)}] The pantry, with its glass doors and open shelves, offers visible storage for frequently accessed or decorative items like glassware and dry goods, while the opaque wooden cabinets provide concealed storage for appliances or less aesthetic bulk supplies.
            \item[(D)] The pantry's transparency is to showcase rarely used, valuable china, contrasting with the wooden cabinets which offer practical, hidden storage for essential everyday tools.
        \end{itemize}
    \end{examplebox}
\end{center}

\clearpage
\twocolumn % Optional: Switch back to two columns if you have more text after the examples
\clearpage
\onecolumn % Switch to full page width to prevent overlap
\tcbset{breakable} % Local force to allow boxes to split across pages

\section{Prompts}
\label{sec:prompts}

\noindent % Ensure the box starts at the left margin
\begin{promptbox}[Scenario Prompt Template]
    You are an assistant tasked with generating image search scenario queries to support visual learning grounded in Bloom's Taxonomy. Your goal is to create scenarios that describe visually rich scenes with minimal or no text, leading to concrete images that can be used for evaluating a model's visual understanding.
    
    \hrulefill
    
    \textbf{Context} \\
    We are creating visual learning prompts based on Bloom's Taxonomy. Each query should describe a clear, specific visual scene that reflects a cognitive ability (Bloom level) applied to a key concept (leaf). The primary goal is to generate scenarios and keywords that produce high-quality, language-independent images from a standard image search engine.
    
    \hrulefill
    
    \textbf{Key Concepts}
    \begin{itemize}
        \item \textbf{Leaf}: The core concept or topic the scenario should visually represent. This is the \textbf{target subject} of the image query. (\textit{Example}: "Pattern Recognition")
        \item \textbf{Bloom Level}: The type of thinking or cognitive ability that the learner is expected to apply to the leaf concept. (\textit{Example}: "Analyzing" $\rightarrow$ discriminating between different parts)
        \item \textbf{Path}: The full hierarchical classification. (\textit{Example}: "Analyzing $\rightarrow$ Differentiating $\rightarrow$ Discriminating $\rightarrow$ Pattern Recognition")
    \end{itemize}

    \hrulefill
    
    \textbf{Bloom's Taxonomy Levels \& Abilities}
    \begin{enumerate}
        \item \textbf{Remembering} – Recall facts and basic concepts.
        \item \textbf{Understanding} – Demonstrate comprehension of meaning.
        \item \textbf{Applying} – Use knowledge in new situations.
        \item \textbf{Analyzing} – Break information into parts.
        \item \textbf{Evaluating} – Make judgments based on criteria.
        \item \textbf{Creating} – Generate new ideas or artifacts.
    \end{enumerate}

    \hrulefill
    
    \textbf{Expected Output (JSON)}
    \begin{verbatim}
    {
        "scenarios": ["Man crossing street while looking at smartphone.", ...],
        "keywords": ["man crossing street looking at phone", ...]
    }
    \end{verbatim}
\end{promptbox}

\vspace{20pt}

\noindent
\begin{promptbox}[VQA Prompt Template]
    You are an expert assistant for creating Visual Question Answering (VQA) benchmarks grounded in Bloom's Taxonomy. Your task is to generate high-quality, insightful question-answer pairs answerable \textit{only} by analyzing the visual content of the image.

    \hrulefill
    
    \textbf{Guidelines}
    \begin{itemize}
        \item \textbf{Image-Grounded}: Both the question and answer MUST be directly derivable from the visual information.
        \item \textbf{Bloom Alignment}: The question must genuinely require the cognitive skill of \texttt{\{\{bloom\_node\}\}}.
        \item \textbf{Deterministic}: Answers should be objective and verifiable.
    \end{itemize}
    
    \hrulefill

    \textbf{Example Output}
    \begin{verbatim}
    [
      {
        "question_en": "Based on the dog's attire, what event is taking place?",
        "answer_en": "A birthday party, indicated by the birthday hat."
      }
    ]
    \end{verbatim}
\end{promptbox}

\vspace{20pt}

\noindent
\begin{promptbox}[Make Multiple Choice Arabic QA Prompt] 
    You are a professional test creator. Your task is to generate a high-quality multiple-choice question in English, and then provide a translation in Modern Standard Arabic.

    \hrulefill
    
    \textbf{Important Guidelines}
    \begin{itemize}
        \item One of the distractors must be a \textbf{trap answer}—a tempting but incorrect option.
        \item Strong trap answers involve confusing similar regional or cultural features (e.g., Qatari vs. Emirati attire).
        \item Provide the correct answer as Choice A in the JSON structure.
    \end{itemize}
\end{promptbox}

\clearpage
\twocolumn % Switch back to two columns if you have more sections
\section{Human Annotation and Quality Control Guidelines}
\label{sec:annotation_guidelines}

Annotators evaluated a stratified subset of the dataset using a custom annotation interface. For each sample, they were provided with the image, the corresponding multiple-choice question in both English and Arabic, the correct answer, and the target taxonomy path. 

Annotators were instructed to verify that all components were accurate, coherent, and meaningful. They were required to either approve the sample as fully valid or flag it under one of the following four specific error categories:

\begin{itemize}
    \item \textbf{Image Quality and Alignment:} Verifying that the image is clear, relevant, and correctly aligns with the targeted Bloom's Taxonomy node.
    \item \textbf{Question Integrity:} Ensuring the English question is natural, unambiguous, grammatically correct, and strictly grounded in both the visual content and the assigned cognitive level.
    \item \textbf{Choice Validity:} Confirming that the multiple-choice options are clear, properly structured, and not misleading, with exactly one unequivocally correct answer.
    \item \textbf{Translation Fidelity:} Evaluating the Arabic translation to ensure it accurately, naturally, and faithfully reflects the semantic meaning of the original English text.
\end{itemize}

Samples that passed all four criteria without issue were marked as valid. If a sample was flagged for any of the above issues, it was recorded for removal or revision. The annotation platform also included concurrency controls to ensure each question was reviewed exactly once by the annotator pool.
\clearpage
\onecolumn % Ensure we are in one-column mode for the wide table
\section{Detailed Results of Benchmarking}
\label{sec:details}

\begin{center}
    \small
    \setlength{\tabcolsep}{3pt} 
    \resizebox{\textwidth}{!}{
        \begin{tabular}{ll *{10}{c}}
        \toprule
        Sub-category & Sub-sub-category & \rotHeader{Gemma-3-4B-en} & \rotHeader{Qwen2-VL-7B-Instruct\_en} & \rotHeader{Qwen2.5-VL-7B-Instruct\_en} & \rotHeader{Gemma-3-27B-it\_en} & \rotHeader{Gemma-3-12B-it\_en} & \rotHeader{Gemma-3-12B-it\_ar} & \rotHeader{Gemma-3-27B-it\_ar} & \rotHeader{Qwen2-VL-7B-Instruct\_ar} & \rotHeader{Qwen2.5-VL-7B-Instruct\_ar} & \rotHeader{Gemma-3-4B-it\_ar} \\
        \midrule
        atypical attribute identification & artistic style & 0.84 & 0.52 & 0.75 & 0.64 & 0.77 & 0.54 & 0.63 & 0.4 & 0.52 & 0.57 \\
        atypical attribute identification & color & 0.75 & 0.56 & 0.88 & 0.18 & 0.57 & 0.53 & 0.59 & 0.38 & 0.65 & 0.47 \\
        atypical attribute identification & shape & 0.62 & 0.44 & 0.64 & 0.48 & 0.53 & 0.52 & 0.55 & 0.34 & 0.48 & 0.41 \\
        atypical attribute identification & size & 0.63 & 0.4 & 0.57 & 0.45 & 0.64 & 0.52 & 0.48 & 0.45 & 0.49 & 0.49 \\
        atypical attribute identification & texture & 0.79 & 0.51 & 0.82 & 0.64 & 0.75 & 0.63 & 0.62 & 0.32 & 0.53 & 0.54 \\
        contextual inference & ambiguity resolution & 0.65 & 0.43 & 0.81 & 0.48 & 0.59 & 0.52 & 0.58 & 0.33 & 0.77 & 0.46 \\
        contextual inference & commonsense reasooning & 0.76 & 0.47 & 0.82 & 0.5 & 0.68 & 0.6 & 0.59 & 0.49 & 0.72 & 0.67 \\
        contextual inference & comparative reasoning & 0.77 & 0.59 & 0.88 & 0.63 & 0.78 & 0.63 & 0.62 & 0.34 & 0.6 & 0.66 \\
        contextual inference & ellipsis resolution & 0.56 & 0.5 & 0.72 & 0.41 & 0.5 & 0.44 & 0.51 & 0.32 & 0.41 & 0.41 \\
        contextual inference & pronoun resolution & 0.51 & 0.41 & 0.72 & 0.32 & 0.46 & 0.49 & 0.56 & 0.31 & 0.49 & 0.45 \\
        logical and scientific reasoning & logic reasoning & 0.68 & 0.53 & 0.71 & 0.51 & 0.6 & 0.62 & 0.6 & 0.34 & 0.59 & 0.62 \\
        logical and scientific reasoning & math reasoning & 0.6 & 0.46 & 0.59 & 0.51 & 0.52 & 0.43 & 0.48 & 0.33 & 0.48 & 0.42 \\
        logical and scientific reasoning & scientific reasoning & 0.74 & 0.56 & 0.81 & 0.47 & 0.62 & 0.6 & 0.62 & 0.42 & 0.63 & 0.56 \\
        structured data analysis & chart analysis & 0.74 & 0.57 & 0.76 & 0.58 & 0.68 & 0.65 & 0.62 & 0.39 & 0.54 & 0.65 \\
        structured data analysis & chemical structure analysis & 0.74 & 0.48 & 0.68 & 0.6 & 0.75 & 0.53 & 0.63 & 0.37 & 0.59 & 0.56 \\
        structured data analysis & diagram analysis & 0.7 & 0.53 & 0.8 & 0.66 & 0.73 & 0.66 & 0.67 & 0.48 & 0.72 & 0.7 \\
        structured data analysis & document analysis & 0.78 & 0.51 & 0.8 & 0.62 & 0.68 & 0.65 & 0.67 & 0.36 & 0.64 & 0.59 \\
        structured data analysis & sheet music analysis & 0.71 & 0.52 & 0.73 & 0.54 & 0.56 & 0.58 & 0.73 & 0.37 & 0.58 & 0.54 \\
        structured data analysis & table analysis & 0.71 & 0.53 & 0.75 & 0.59 & 0.67 & 0.63 & 0.65 & 0.39 & 0.56 & 0.55 \\
        \bottomrule
        \end{tabular}
    }
    \captionof{table}{Likelihood-based Scoring (LBS) performance comparison across models for category: \textit{analyze}}
    \label{tab:analyze}
\end{center}

\begin{table*}[h]
\centering
\small
\setlength{\tabcolsep}{4pt}
\resizebox{\textwidth}{!}{
\begin{tabular}{llccccccccccc}
\toprule
Sub-category & Sub-sub-category & \rotHeader{Gemma-3-4B-en} & \rotHeader{Qwen2-VL-7B-Instruct\_en} & \rotHeader{Qwen2.5-VL-7B-Instruct\_en} & \rotHeader{Gemma-3-27B-it\_en} & \rotHeader{Gemma-3-12B-it\_en} & \rotHeader{Gemma-3-12B-it\_ar} & \rotHeader{Gemma-3-27B-it\_ar} & \rotHeader{Qwen2-VL-7B-Instruct\_ar} & \rotHeader{Qwen2.5-VL-7B-Instruct\_ar} & \rotHeader{Gemma-3-4B-it\_ar} \\
\midrule
basic logic operation & coordination interpretation & 0.71 & 0.55 & 0.84 & 0.6 & 0.53 & 0.5 & 0.56 & 0.31 & 0.47 & 0.56 \\
basic logic operation & negation understanding & 0.4 & 0.38 & 0.57 & 0.27 & 0.32 & 0.24 & 0.29 & 0.27 & 0.32 & 0.24 \\
basic logic operation & word order understanding & 0.34 & 0.24 & 0.48 & 0.13 & 0.24 & 0.21 & 0.24 & 0.21 & 0.37 & 0.24 \\
knowledge application & applying a design principle & 0.64 & 0.58 & 0.74 & 0.43 & 0.53 & 0.45 & 0.52 & 0.39 & 0.55 & 0.49 \\
knowledge application & applying a mathematical formula & 0.39 & 0.38 & 0.51 & 0.28 & 0.26 & 0.36 & 0.26 & 0.28 & 0.41 & 0.26 \\
knowledge application & applying a scientific concept & 0.82 & 0.64 & 0.79 & 0.6 & 0.6 & 0.55 & 0.63 & 0.41 & 0.58 & 0.59 \\
knowledge application & procedural step following & 0.65 & 0.53 & 0.72 & 0.43 & 0.44 & 0.44 & 0.56 & 0.39 & 0.52 & 0.45 \\
\bottomrule
\end{tabular}
}
\caption{Likelihood-based Scoring (LBS) performance comparison across models for category: \textit{apply}}
\label{tab:apply}
\end{table*}

\begin{table*}[t]
\centering
\small
\setlength{\tabcolsep}{4pt}
\resizebox{\textwidth}{!}{
\begin{tabular}{llccccccccccc}
\toprule
Sub-category & Sub-sub-category & \rotHeader{Gemma-3-4B-en} & \rotHeader{Qwen2-VL-7B-Instruct\_en} & \rotHeader{Qwen2.5-VL-7B-Instruct\_en} & \rotHeader{Gemma-3-27B-it\_en} & \rotHeader{Gemma-3-12B-it\_en} & \rotHeader{Gemma-3-12B-it\_ar} & \rotHeader{Gemma-3-27B-it\_ar} & \rotHeader{Qwen2-VL-7B-Instruct\_ar} & \rotHeader{Qwen2.5-VL-7B-Instruct\_ar} & \rotHeader{Gemma-3-4B-it\_ar} \\
\midrule
creative generation & creative title generation & 0.41 & 0.38 & 0.56 & 0.31 & 0.35 & 0.25 & 0.31 & 0.26 & 0.43 & 0.28 \\
creative generation & image captioning & 0.64 & 0.45 & 0.74 & 0.38 & 0.49 & 0.39 & 0.49 & 0.38 & 0.52 & 0.34 \\
creative generation & joke & 0.31 & 0.33 & 0.39 & 0.39 & 0.3 & 0.2 & 0.3 & 0.15 & 0.2 & 0.17 \\
creative generation & meme caption creative generation & 0.47 & 0.33 & 0.47 & 0.48 & 0.5 & 0.36 & 0.43 & 0.21 & 0.31 & 0.45 \\
creative generation & poem & 0.45 & 0.38 & 0.62 & 0.28 & 0.42 & 0.19 & 0.25 & 0.21 & 0.26 & 0.23 \\
creative generation & short story & 0.7 & 0.47 & 0.63 & 0.51 & 0.7 & 0.6 & 0.63 & 0.33 & 0.65 & 0.51 \\
creative generation & visual storytelling & 0.7 & 0.48 & 0.72 & 0.56 & 0.64 & 0.5 & 0.52 & 0.27 & 0.41 & 0.41 \\
structured creation & counterfactual creation & 0.66 & 0.51 & 0.7 & 0.51 & 0.59 & 0.46 & 0.51 & 0.36 & 0.42 & 0.39 \\
structured creation & designing an experiment & 0.69 & 0.55 & 0.61 & 0.65 & 0.71 & 0.58 & 0.55 & 0.44 & 0.53 & 0.6 \\
structured creation & dialogue generation & 0.37 & 0.39 & 0.54 & 0.34 & 0.46 & 0.22 & 0.27 & 0.27 & 0.34 & 0.24 \\
structured creation & image-based question generation & 0.63 & 0.45 & 0.73 & 0.3 & 0.4 & 0.3 & 0.25 & 0.42 & 0.48 & 0.33 \\
\bottomrule
\end{tabular}
}
\caption{Likelihood-based Scoring (LBS) performance comparison across models for category: \textit{create}}
\label{tab:create}
\end{table*}

\begin{table*}[t]
\centering
\small
\setlength{\tabcolsep}{4pt}
\resizebox{\textwidth}{!}{
\begin{tabular}{llcccccccccc}
\toprule
Sub-category & Sub-sub-category & \rotHeader{Gemma-3-4B-en} & \rotHeader{Qwen2-VL-7B-Instruct\_en} & \rotHeader{Qwen2.5-VL-7B-Instruct\_en} & \rotHeader{Gemma-3-27B-it\_en} & \rotHeader{Gemma-3-12B-it\_en} & \rotHeader{Gemma-3-12B-it\_ar} & \rotHeader{Gemma-3-27B-it\_ar} & \rotHeader{Qwen2-VL-7B-Instruct\_ar} & \rotHeader{Qwen2.5-VL-7B-Instruct\_ar} & \rotHeader{Gemma-3-4B-it\_ar} \\
\midrule
harm-safety evaluation & age-appropriateness & 0.75 & 0.49 & 0.79 & 0.38 & 0.59 & 0.49 & 0.52 & 0.44 & 0.71 & 0.44 \\
harm-safety evaluation & contextual suitability & 0.72 & 0.54 & 0.86 & 0.41 & 0.63 & 0.51 & 0.5 & 0.37 & 0.54 & 0.44 \\
harm-safety evaluation &  cultural sensitivity &  0.71 & 0.47 & 0.88 & 0.32 & 0.59 & 0.56 & 0.47 & 0.35 & 0.59 & 0.65 \\
harm-safety evaluation & safety evaluation & 0.75 & 0.44 & 0.8 & 0.4 & 0.56 & 0.49 & 0.53 & 0.39 & 0.47 & 0.47 \\
harm-safety evaluation & toxicity detection & 0.68 & 0.39 & 0.7 & 0.42 & 0.58 & 0.65 & 0.61 & 0.4 & 0.54 & 0.64 \\
logical coherence evaluation & conflicting scenario evaluation & 0.66 & 0.46 & 0.56 & 0.32 & 0.58 & 0.46 & 0.51 & 0.37 & 0.53 & 0.46 \\
logical coherence evaluation & object hallucination evaluation & 0.61 & 0.5 & 0.76 & 0.33 & 0.49 & 0.41 & 0.51 & 0.42 & 0.57 & 0.41 \\
quality evaluation & artistic evaluation & 0.76 & 0.5 & 0.82 & 0.53 & 0.59 & 0.51 & 0.47 & 0.28 & 0.55 & 0.46 \\
quality evaluation & image quality assessment & 0.63 & 0.5 & 0.78 & 0.39 & 0.66 & 0.52 & 0.54 & 0.28 & 0.57 & 0.52 \\
\bottomrule
\end{tabular}
}
\caption{Likelihood-based Scoring (LBS) performance comparison across models for category: \textit{evaluate}}
\label{tab:evaluate}
\end{table*}

\begin{table*}[t]
\centering
\small
\setlength{\tabcolsep}{4pt}
\resizebox{\textwidth}{!}{
\begin{tabular}{llcccccccccc}
\toprule
Sub-category & Sub-sub-category & \rotHeader{Gemma-3-4B-en} & \rotHeader{Qwen2-VL-7B-Instruct\_en} & \rotHeader{Qwen2.5-VL-7B-Instruct\_en} & \rotHeader{Gemma-3-27B-it\_en} & \rotHeader{Gemma-3-12B-it\_en} & \rotHeader{Gemma-3-12B-it\_ar} & \rotHeader{Gemma-3-27B-it\_ar} & \rotHeader{Qwen2-VL-7B-Instruct\_ar} & \rotHeader{Qwen2.5-VL-7B-Instruct\_ar} & \rotHeader{Gemma-3-4B-it\_ar} \\
\midrule
activity recognition & individual activities & 0.62 & 0.48 & 0.75 & 0.25 & 0.3 & 0.34 & 0.46 & 0.18 & 0.62 & 0.34 \\
activity recognition & interactions & 0.6 & 0.46 & 0.65 & 0.3 & 0.4 & 0.37 & 0.44 & 0.37 & 0.51 & 0.3 \\
activity recognition & professions & 0.55 & 0.39 & 0.73 & 0.27 & 0.36 & 0.23 & 0.32 & 0.27 & 0.68 & 0.23 \\
attribute recognition & artistic style & 0.48 & 0.47 & 0.66 & 0.19 & 0.34 & 0.3 & 0.26 & 0.32 & 0.51 & 0.21 \\
attribute recognition & color & 0.56 & 0.25 & 0.33 & 0.07 & 0.21 & 0.25 & 0.24 & 0.25 & 0.19 & 0.37 \\
attribute recognition & shape & 0.48 & 0.24 & 0.31 & 0.18 & 0.27 & 0.17 & 0.17 & 0.27 & 0.29 & 0.19 \\
attribute recognition & size & 0.35 & 0.27 & 0.53 & 0.15 & 0.11 & 0.16 & 0.2 & 0.21 & 0.35 & 0.15 \\
attribute recognition & texture & 0.4 & 0.38 & 0.59 & 0.17 & 0.21 & 0.3 & 0.34 & 0.3 & 0.41 & 0.3 \\
core object recognition & animals & 0.52 & 0.21 & 0.34 & 0.06 & 0.18 & 0.28 & 0.31 & 0.21 & 0.26 & 0.29 \\
core object recognition & artifacts & 0.46 & 0.29 & 0.48 & 0.17 & 0.28 & 0.32 & 0.31 & 0.37 & 0.49 & 0.35 \\
core object recognition & arts & 0.41 & 0.36 & 0.57 & 0.2 & 0.28 & 0.2 & 0.29 & 0.3 & 0.47 & 0.22 \\
core object recognition & clothing-accessories & 0.46 & 0.27 & 0.35 & 0.09 & 0.2 & 0.27 & 0.35 & 0.22 & 0.34 & 0.35 \\
core object recognition & common objects & 0.47 & 0.11 & 0.36 & 0.14 & 0.19 & 0.19 & 0.2 & 0.18 & 0.34 & 0.19 \\
core object recognition & food-beverage & 0.58 & 0.16 & 0.44 & 0.11 & 0.23 & 0.18 & 0.25 & 0.23 & 0.41 & 0.16 \\
core object recognition & indoor scenes & 0.49 & 0.25 & 0.59 & 0.17 & 0.28 & 0.22 & 0.26 & 0.25 & 0.46 & 0.28 \\
core object recognition & outdoor scenes & 0.49 & 0.11 & 0.55 & 0.11 & 0.19 & 0.22 & 0.24 & 0.24 & 0.45 & 0.17 \\
core object recognition & people & 0.44 & 0.19 & 0.34 & 0.1 & 0.13 & 0.12 & 0.27 & 0.22 & 0.31 & 0.22 \\
core object recognition & produce-plants & 0.64 & 0.22 & 0.37 & 0.2 & 0.23 & 0.18 & 0.2 & 0.23 & 0.34 & 0.15 \\
core object recognition & technology-electronics & 0.43 & 0.26 & 0.45 & 0.11 & 0.26 & 0.22 & 0.32 & 0.3 & 0.46 & 0.26 \\
core object recognition & vehicles & 0.51 & 0.18 & 0.45 & 0.16 & 0.22 & 0.21 & 0.29 & 0.26 & 0.49 & 0.26 \\
symbol recognition & app-tech icons & 0.52 & 0.23 & 0.62 & 0.1 & 0.23 & 0.33 & 0.33 & 0.27 & 0.44 & 0.38 \\
symbol recognition & astrological-zodiac signs & 0.45 & 0.26 & 0.29 & 0.03 & 0.11 & 0.16 & 0.26 & 0.16 & 0.18 & 0.24 \\
symbol recognition & currency symbols & 0.67 & 0.29 & 0.61 & 0.14 & 0.31 & 0.24 & 0.29 & 0.25 & 0.47 & 0.39 \\
symbol recognition & emoji & 0.51 & 0.23 & 0.54 & 0.17 & 0.19 & 0.23 & 0.35 & 0.35 & 0.48 & 0.33 \\
symbol recognition & flags & 0.49 & 0.27 & 0.45 & 0.12 & 0.27 & 0.2 & 0.2 & 0.24 & 0.28 & 0.3 \\
symbol recognition & formula & 0.34 & 0.34 & 0.5 & 0.16 & 0.15 & 0.18 & 0.27 & 0.23 & 0.32 & 0.35 \\
symbol recognition & logos-brands & 0.58 & 0.28 & 0.42 & 0.13 & 0.21 & 0.25 & 0.32 & 0.26 & 0.53 & 0.32 \\
symbol recognition & music & 0.47 & 0.24 & 0.48 & 0.15 & 0.27 & 0.18 & 0.23 & 0.28 & 0.52 & 0.18 \\
symbol recognition & religious symbols & 0.47 & 0.32 & 0.51 & 0.16 & 0.21 & 0.18 & 0.26 & 0.19 & 0.35 & 0.21 \\
symbol recognition & safety symbols & 0.58 & 0.27 & 0.59 & 0.04 & 0.19 & 0.15 & 0.23 & 0.21 & 0.45 & 0.29 \\
symbol recognition & traffic signs & 0.54 & 0.27 & 0.42 & 0.12 & 0.19 & 0.16 & 0.22 & 0.3 & 0.32 & 0.34 \\
text attribute recognition & books & 0.41 & 0.21 & 0.34 & 0.19 & 0.2 & 0.29 & 0.26 & 0.25 & 0.4 & 0.29 \\
text attribute recognition & documents & 0.5 & 0.36 & 0.61 & 0.2 & 0.21 & 0.2 & 0.26 & 0.32 & 0.45 & 0.2 \\
text attribute recognition & handwriting & 0.41 & 0.27 & 0.33 & 0.16 & 0.33 & 0.28 & 0.29 & 0.27 & 0.35 & 0.32 \\
text attribute recognition & lines & 0.39 & 0.2 & 0.43 & 0.17 & 0.17 & 0.23 & 0.35 & 0.29 & 0.33 & 0.27 \\
text attribute recognition & newsletter & 0.37 & 0.25 & 0.6 & 0.16 & 0.17 & 0.21 & 0.17 & 0.38 & 0.57 & 0.3 \\
text attribute recognition & number & 0.48 & 0.18 & 0.38 & 0.03 & 0.14 & 0.14 & 0.19 & 0.22 & 0.29 & 0.34 \\
text attribute recognition & power point slides & 0.49 & 0.32 & 0.49 & 0.14 & 0.25 & 0.16 & 0.22 & 0.35 & 0.46 & 0.21 \\
text attribute recognition & scene text & 0.45 & 0.18 & 0.38 & 0.08 & 0.26 & 0.19 & 0.25 & 0.36 & 0.41 & 0.3 \\
\bottomrule
\end{tabular}
}
\caption{Likelihood-based Scoring (LBS) performance comparison across models for category: \textit{remember}}
\label{tab:remember}
\end{table*}

\begin{table*}[t]
\centering
\small
\setlength{\tabcolsep}{4pt}
\resizebox{\textwidth}{!}{
\begin{tabular}{llcccccccccc}
\toprule
Sub-category & Sub-sub-category & \rotHeader{Gemma-3-4B-en} & \rotHeader{Qwen2-VL-7B-Instruct\_en} & \rotHeader{Qwen2.5-VL-7B-Instruct\_en} & \rotHeader{Gemma-3-27B-it\_en} & \rotHeader{Gemma-3-12B-it\_en} & \rotHeader{Gemma-3-12B-it\_ar} & \rotHeader{Gemma-3-27B-it\_ar} & \rotHeader{Qwen2-VL-7B-Instruct\_ar} & \rotHeader{Qwen2.5-VL-7B-Instruct\_ar} & \rotHeader{Gemma-3-4B-it\_ar} \\
\midrule
cognitive understanding & facial-emotional understanding & 0.65 & 0.52 & 0.85 & 0.41 & 0.55 & 0.52 & 0.63 & 0.37 & 0.65 & 0.48 \\
cognitive understanding & lingual expression alternation & 0.55 & 0.32 & 0.45 & 0.19 & 0.23 & 0.31 & 0.31 & 0.34 & 0.45 & 0.3 \\
cognitive understanding & semantic understanding (knowledge) & 0.77 & 0.59 & 0.93 & 0.38 & 0.55 & 0.57 & 0.65 & 0.45 & 0.8 & 0.64 \\
cognitive understanding & visual alternation & 0.75 & 0.56 & 0.87 & 0.46 & 0.62 & 0.61 & 0.48 & 0.44 & 0.73 & 0.48 \\
compositional attribute recognition & artistic style & 0.8 & 0.65 & 0.91 & 0.53 & 0.73 & 0.65 & 0.68 & 0.41 & 0.71 & 0.61 \\
compositional attribute recognition & shape & 0.67 & 0.51 & 0.73 & 0.49 & 0.62 & 0.49 & 0.49 & 0.4 & 0.54 & 0.49 \\
compositional attribute recognition & size & 0.66 & 0.59 & 0.86 & 0.3 & 0.61 & 0.47 & 0.53 & 0.44 & 0.46 & 0.4 \\
compositional attribute recognition & texture & 0.88 & 0.58 & 0.87 & 0.64 & 0.76 & 0.7 & 0.65 & 0.23 & 0.52 & 0.52 \\
compositional core object recognition & animals & 0.78 & 0.68 & 0.94 & 0.41 & 0.63 & 0.53 & 0.62 & 0.4 & 0.59 & 0.54 \\
compositional core object recognition & artifacts & 0.82 & 0.64 & 0.88 & 0.55 & 0.72 & 0.61 & 0.61 & 0.47 & 0.59 & 0.54 \\
compositional core object recognition & arts & 0.83 & 0.61 & 0.93 & 0.46 & 0.67 & 0.49 & 0.57 & 0.4 & 0.7 & 0.52 \\
compositional core object recognition & closed vocabulary object detection & 0.71 & 0.48 & 0.76 & 0.45 & 0.64 & 0.57 & 0.66 & 0.37 & 0.62 & 0.55 \\
compositional core object recognition & clothing-accessories & 0.83 & 0.63 & 0.94 & 0.39 & 0.65 & 0.54 & 0.54 & 0.42 & 0.7 & 0.6 \\
compositional core object recognition & common objects & 0.81 & 0.61 & 0.9 & 0.5 & 0.6 & 0.6 & 0.65 & 0.31 & 0.58 & 0.56 \\
compositional core object recognition & food-beverage & 0.82 & 0.72 & 0.93 & 0.59 & 0.69 & 0.53 & 0.58 & 0.31 & 0.66 & 0.52 \\
compositional core object recognition & indoor scenes & 0.76 & 0.67 & 0.98 & 0.46 & 0.68 & 0.62 & 0.75 & 0.44 & 0.84 & 0.57 \\
compositional core object recognition & outdoor scenes & 0.82 & 0.69 & 0.96 & 0.46 & 0.8 & 0.64 & 0.72 & 0.52 & 0.76 & 0.63 \\
compositional core object recognition & people & 0.77 & 0.63 & 0.96 & 0.47 & 0.62 & 0.51 & 0.59 & 0.45 & 0.84 & 0.62 \\
compositional core object recognition & produce-plants & 0.69 & 0.6 & 0.87 & 0.4 & 0.6 & 0.57 & 0.57 & 0.28 & 0.57 & 0.51 \\
compositional core object recognition & technology-electronics & 0.76 & 0.55 & 0.85 & 0.44 & 0.69 & 0.53 & 0.62 & 0.38 & 0.75 & 0.62 \\
compositional core object recognition & vehicles & 0.76 & 0.6 & 0.92 & 0.37 & 0.56 & 0.48 & 0.57 & 0.39 & 0.73 & 0.52 \\
\bottomrule
\end{tabular}
}
\caption{Likelihood-based Scoring (LBS) performance comparison across models for category: \textit{understand}}
\label{tab:understand}
\end{table*}

%%%%%%%%%%%%%%%%%%%%%%%%%%%%%%%%%%%%%%%%%%%
%%%%%%%%%%%%%%%%%%%%%%%%%%%%%%%%%%%%%%%%%%%
\begin{table*}[t]
\centering
\small
\setlength{\tabcolsep}{4pt}
\resizebox{\textwidth}{!}{
\begin{tabular}{llcccccccccccccc}
\toprule
Sub-category & Sub-sub-category & \rotHeader{Gemma-3-4B-it\_en} & \rotHeader{Qwen2-VL-7B-Instruct\_en} & \rotHeader{Qwen2.5-VL-7B-Instruct\_en} & \rotHeader{Gemma-3-27b-it\_en} & \rotHeader{Gemma-3-12b-it\_en} & \rotHeader{GPT4omini\_en} & \rotHeader{Gemma-3-12B-it\_ar} & \rotHeader{Gemma-3-27B-it\_ar} & \rotHeader{GPT4omini\_ar} & \rotHeader{Qwen2-VL-7B-Instruct\_ar} & \rotHeader{Qwen2.5-VL-7B-Instruct\_ar} & \rotHeader{Gemma-3-4B-it\_ar} \\
\midrule
atypical attribute identification & artistic style & 0.91 & 0.93 & 0.94 & 0.98 & 0.94 & 0.86 & 0.93 & 0.95 & 0.85 & 0.83 & 0.8 & 0.85 \\
atypical attribute identification & color & 0.9 & 0.94 & 0.96 & 0.97 & 0.94 & 0.94 & 0.87 & 0.96 & 0.87 & 0.91 & 0.88 & 0.81 \\
atypical attribute identification & shape & 0.71 & 0.79 & 0.82 & 0.9 & 0.82 & 0.81 & 0.74 & 0.79 & 0.66 & 0.67 & 0.64 & 0.6 \\
atypical attribute identification & size & 0.79 & 0.82 & 0.81 & 0.84 & 0.79 & 0.78 & 0.84 & 0.82 & 0.69 & 0.7 & 0.7 & 0.7 \\
atypical attribute identification & texture & 0.86 & 0.8 & 0.84 & 0.89 & 0.91 & 0.82 & 0.83 & 0.84 & 0.7 & 0.72 & 0.76 & 0.78 \\
contextual inference & ambiguity resolution & 0.9 & 0.94 & 0.97 & 0.96 & 0.99 & 0.9 & 0.93 & 0.99 & 0.88 & 0.78 & 0.87 & 0.83 \\
contextual inference & commonsense reasooning & 0.82 & 0.95 & 0.86 & 0.91 & 0.92 & 0.87 & 0.9 & 0.91 & 0.83 & 0.88 & 0.88 & 0.79 \\
contextual inference & comparative reasoning & 0.89 & 0.93 & 0.98 & 0.89 & 0.91 & 0.88 & 0.89 & 0.88 & 0.73 & 0.84 & 0.89 & 0.83 \\
contextual inference & ellipsis resolution & 0.84 & 0.9 & 0.91 & 0.94 & 0.96 & 0.91 & 0.93 & 0.9 & 0.93 & 0.84 & 0.9 & 0.76 \\
contextual inference & pronoun resolution & 0.92 & 0.92 & 0.96 & 0.94 & 0.9 & 0.89 & 0.9 & 0.89 & 0.83 & 0.8 & 0.76 & 0.82 \\
logical and scientific reasoning & logic reasoning & 0.75 & 0.77 & 0.77 & 0.9 & 0.79 & 0.83 & 0.81 & 0.87 & 0.79 & 0.69 & 0.71 & 0.67 \\
logical and scientific reasoning & math reasoning & 0.62 & 0.73 & 0.73 & 0.64 & 0.68 & 0.67 & 0.65 & 0.65 & 0.65 & 0.72 & 0.63 & 0.56 \\
logical and scientific reasoning & scientific reasoning & 0.84 & 0.9 & 0.85 & 0.87 & 0.91 & 0.9 & 0.85 & 0.93 & 0.84 & 0.81 & 0.83 & 0.74 \\
structured data analysis & chart analysis & 0.81 & 0.84 & 0.87 & 0.86 & 0.82 & 0.76 & 0.84 & 0.83 & 0.72 & 0.8 & 0.82 & 0.68 \\
structured data analysis & chemical structure analysis & 0.79 & 0.71 & 0.77 & 0.88 & 0.82 & 0.74 & 0.86 & 0.88 & 0.7 & 0.68 & 0.75 & 0.73 \\
structured data analysis & diagram analysis & 0.83 & 0.89 & 0.86 & 0.89 & 0.92 & 0.84 & 0.91 & 0.89 & 0.81 & 0.78 & 0.84 & 0.72 \\
structured data analysis & document analysis & 0.86 & 0.9 & 0.97 & 0.92 & 0.97 & 0.91 & 0.9 & 0.92 & 0.9 & 0.88 & 0.91 & 0.83 \\
structured data analysis & sheet music analysis & 0.71 & 0.83 & 0.85 & 0.9 & 0.87 & 0.77 & 0.85 & 0.87 & 0.81 & 0.77 & 0.81 & 0.63 \\
structured data analysis & table analysis & 0.8 & 0.87 & 0.85 & 0.89 & 0.93 & 0.87 & 0.88 & 0.88 & 0.76 & 0.77 & 0.76 & 0.72 \\
\bottomrule
\end{tabular}
}
\caption{Regex Based Extraction (RAE) Performance comparison across models for category: \textit{analyze}}
\label{tab:analyze_reg}
\end{table*}

\begin{table*}[t]
\centering
\small
\setlength{\tabcolsep}{4pt}
\resizebox{\textwidth}{!}{
\begin{tabular}{llcccccccccccccc}
\toprule
Sub-category & Sub-sub-category & \rotHeader{Gemma-3-4B-it\_en} & \rotHeader{Qwen2-VL-7B-Instruct\_en} & \rotHeader{Qwen2.5-VL-7B-Instruct\_en} & \rotHeader{Gemma-3-27b-it\_en} & \rotHeader{Gemma-3-12b-it\_en} & \rotHeader{GPT4omini\_en} & \rotHeader{Gemma-3-12B-it\_ar} & \rotHeader{Gemma-3-27B-it\_ar} & \rotHeader{GPT4omini\_ar} & \rotHeader{Qwen2-VL-7B-Instruct\_ar} & \rotHeader{Qwen2.5-VL-7B-Instruct\_ar} & \rotHeader{Gemma-3-4B-it\_ar} \\
\midrule
basic logic operation & coordination interpretation & 0.87 & 0.89 & 0.94 & 0.89 & 0.92 & 0.87 & 0.89 & 0.9 & 0.74 & 0.79 & 0.85 & 0.77 \\
basic logic operation & negation understanding & 0.7 & 0.75 & 0.77 & 0.88 & 0.79 & 0.74 & 0.79 & 0.82 & 0.69 & 0.64 & 0.65 & 0.58 \\
basic logic operation & word order understanding & 0.66 & 0.67 & 0.78 & 0.76 & 0.75 & 0.52 & 0.66 & 0.79 & 0.49 & 0.61 & 0.52 & 0.58 \\
knowledge application & applying a design principle & 0.79 & 0.9 & 0.9 & 0.86 & 0.88 & 0.88 & 0.87 & 0.83 & 0.81 & 0.75 & 0.75 & 0.75 \\
knowledge application & applying a mathematical formula & 0.51 & 0.66 & 0.75 & 0.8 & 0.72 & 0.75 & 0.64 & 0.8 & 0.67 & 0.59 & 0.69 & 0.44 \\
knowledge application & applying a scientific concept & 0.74 & 0.89 & 0.89 & 0.92 & 0.85 & 0.86 & 0.92 & 0.9 & 0.81 & 0.75 & 0.74 & 0.75 \\
knowledge application & procedural step following & 0.79 & 0.81 & 0.73 & 0.85 & 0.77 & 0.79 & 0.81 & 0.83 & 0.76 & 0.71 & 0.71 & 0.6 \\
\bottomrule
\end{tabular}
}
\caption{Regex Based Extraction (RAE) Performance comparison across models for category: \textit{apply}}
\label{tab:apply_reg}
\end{table*}

\begin{table*}[t]
\centering
\small
\setlength{\tabcolsep}{4pt}
\resizebox{\textwidth}{!}{
\begin{tabular}{llcccccccccccccc}
\toprule
Sub-category & Sub-sub-category & \rotHeader{Gemma-3-4B-it\_en} & \rotHeader{Qwen2-VL-7B-Instruct\_en} & \rotHeader{Qwen2.5-VL-7B-Instruct\_en} & \rotHeader{Gemma-3-27b-it\_en} & \rotHeader{Gemma-3-12b-it\_en} & \rotHeader{GPT4omini\_en} & \rotHeader{Gemma-3-12B-it\_ar} & \rotHeader{Gemma-3-27B-it\_ar} & \rotHeader{GPT4omini\_ar} & \rotHeader{Qwen2-VL-7B-Instruct\_ar} & \rotHeader{Qwen2.5-VL-7B-Instruct\_ar} & \rotHeader{Gemma-3-4B-it\_ar} \\
\midrule
creative generation & creative title generation & 0.64 & 0.77 & 0.84 & 0.77 & 0.74 & 0.78 & 0.68 & 0.64 & 0.64 & 0.58 & 0.62 & 0.62 \\
creative generation & image captioning & 0.77 & 0.75 & 0.79 & 0.83 & 0.81 & 0.73 & 0.78 & 0.83 & 0.74 & 0.69 & 0.71 & 0.7 \\
creative generation & joke & 0.54 & 0.5 & 0.43 & 0.69 & 0.56 & 0.59 & 0.56 & 0.63 & 0.52 & 0.39 & 0.48 & 0.44 \\
creative generation & meme caption creative generation & 0.53 & 0.53 & 0.6 & 0.74 & 0.62 & 0.6 & 0.55 & 0.69 & 0.5 & 0.45 & 0.53 & 0.47 \\
creative generation & poem & 0.85 & 0.79 & 0.91 & 0.91 & 0.92 & 0.72 & 0.83 & 0.81 & 0.55 & 0.64 & 0.74 & 0.7 \\
creative generation & short story & 0.79 & 0.74 & 0.91 & 0.91 & 0.86 & 0.88 & 0.88 & 0.86 & 0.88 & 0.72 & 0.72 & 0.72 \\
creative generation & visual storytelling & 0.86 & 0.81 & 0.86 & 0.84 & 0.91 & 0.75 & 0.84 & 0.78 & 0.78 & 0.81 & 0.88 & 0.77 \\
structured creation & counterfactual creation & 0.72 & 0.73 & 0.72 & 0.84 & 0.81 & 0.73 & 0.72 & 0.73 & 0.66 & 0.68 & 0.58 & 0.58 \\
structured creation & designing an experiment & 0.84 & 0.89 & 0.94 & 0.95 & 0.87 & 0.94 & 0.84 & 0.92 & 0.89 & 0.82 & 0.84 & 0.76 \\
structured creation & dialogue generation & 0.76 & 0.8 & 0.85 & 0.86 & 0.86 & 0.88 & 0.78 & 0.83 & 0.78 & 0.75 & 0.71 & 0.66 \\
structured creation & image-based question generation & 0.82 & 0.87 & 0.88 & 0.9 & 0.88 & 0.85 & 0.83 & 0.87 & 0.72 & 0.8 & 0.82 & 0.77 \\
\bottomrule
\end{tabular}
}
\caption{Regex Based Extraction (RAE) Performance comparison across models for category: \textit{create}}
\label{tab:create_reg}
\end{table*}

\begin{table*}[t]
\centering
\small
\setlength{\tabcolsep}{4pt}
\resizebox{\textwidth}{!}{
\begin{tabular}{llcccccccccccccc}
\toprule
Sub-category & Sub-sub-category & \rotHeader{Gemma-3-4B-it\_en} & \rotHeader{Qwen2-VL-7B-Instruct\_en} & \rotHeader{Qwen2.5-VL-7B-Instruct\_en} & \rotHeader{Gemma-3-27b-it\_en} & \rotHeader{Gemma-3-12b-it\_en} & \rotHeader{GPT4omini\_en} & \rotHeader{Gemma-3-12B-it\_ar} & \rotHeader{Gemma-3-27B-it\_ar} & \rotHeader{GPT4omini\_ar} & \rotHeader{Qwen2-VL-7B-Instruct\_ar} & \rotHeader{Qwen2.5-VL-7B-Instruct\_ar} & \rotHeader{Gemma-3-4B-it\_ar}\\
\midrule
harm-safety evaluation & age-appropriateness & 0.89 & 0.95 & 0.92 & 0.95 & 0.97 & 0.92 & 0.94 & 0.92 & 0.87 & 0.87 & 0.86 & 0.84 \\
harm-safety evaluation & contextual suitability & 0.79 & 0.91 & 0.89 & 0.99 & 0.93 & 0.97 & 0.97 & 0.97 & 0.87 & 0.73 & 0.81 & 0.71 \\
harm-safety evaluation & cultural sensitivity & 0.82 & 1.0 & 0.97 & 0.94 & 1.0 & 1.0 & 0.94 & 0.91 & 0.91 & 0.91 & 0.91 & 0.79 \\
harm-safety evaluation & safety evaluation & 0.83 & 0.92 & 0.96 & 0.93 & 0.91 & 0.92 & 0.92 & 0.93 & 0.92 & 0.84 & 0.88 & 0.8 \\
harm-safety evaluation & toxicity detection & 0.82 & 0.82 & 0.88 & 0.88 & 0.93 & 0.79 & 0.86 & 0.89 & 0.74 & 0.77 & 0.82 & 0.75 \\
logical coherence evaluation & conflicting scenario evaluation & 0.63 & 0.75 & 0.73 & 0.8 & 0.78 & 0.73 & 0.68 & 0.76 & 0.66 & 0.49 & 0.58 & 0.47 \\
logical coherence evaluation & object hallucination evaluation & 0.62 & 0.82 & 0.84 & 0.86 & 0.87 & 0.74 & 0.88 & 0.83 & 0.79 & 0.74 & 0.71 & 0.59 \\
quality evaluation & artistic evaluation & 0.81 & 0.95 & 0.94 & 0.95 & 0.92 & 0.91 & 0.95 & 0.91 & 0.9 & 0.88 & 0.88 & 0.86 \\
quality evaluation & image quality assessment & 0.67 & 0.88 & 0.9 & 0.84 & 0.8 & 0.79 & 0.82 & 0.87 & 0.78 & 0.87 & 0.89 & 0.74 \\
\bottomrule
\end{tabular}
}
\caption{Regex Based Extraction (RAE) Performance comparison across models for category: \textit{evaluate}}
\label{tab:evaluate_reg}
\end{table*}

\begin{table*}[t]
\centering
\small
\setlength{\tabcolsep}{4pt}
\resizebox{\textwidth}{!}{
\begin{tabular}{llcccccccccccccc}
\toprule
Sub-category & Sub-sub-category & \rotHeader{Gemma-3-4B-it\_en} & \rotHeader{Qwen2-VL-7B-Instruct\_en} & \rotHeader{Qwen2.5-VL-7B-Instruct\_en} & \rotHeader{Gemma-3-27b-it\_en} & \rotHeader{Gemma-3-12b-it\_en} & \rotHeader{GPT4omini\_en} & \rotHeader{Gemma-3-12B-it\_ar} & \rotHeader{Gemma-3-27B-it\_ar} & \rotHeader{GPT4omini\_ar} & \rotHeader{Qwen2-VL-7B-Instruct\_ar} & \rotHeader{Qwen2.5-VL-7B-Instruct\_ar} & \rotHeader{Gemma-3-4B-it\_ar} \\
\midrule
activity recognition & individual activities & 0.87 & 0.87 & 0.95 & 0.93 & 0.92 & 0.89 & 0.95 & 0.98 & 0.9 & 0.89 & 0.92 & 0.89 \\
activity recognition & interactions & 0.84 & 0.86 & 0.91 & 0.84 & 0.86 & 0.86 & 0.88 & 0.88 & 0.82 & 0.84 & 0.88 & 0.79 \\
activity recognition & professions & 0.95 & 0.96 & 0.95 & 0.95 & 0.96 & 0.96 & 0.91 & 0.93 & 0.93 & 0.95 & 0.95 & 0.89 \\
attribute recognition & artistic style & 0.83 & 0.88 & 0.88 & 0.92 & 0.94 & 0.9 & 0.9 & 0.92 & 0.82 & 0.83 & 0.82 & 0.77 \\
attribute recognition & color & 0.59 & 0.78 & 0.77 & 0.78 & 0.73 & 0.69 & 0.65 & 0.75 & 0.59 & 0.66 & 0.69 & 0.52 \\
attribute recognition & shape & 0.64 & 0.8 & 0.71 & 0.77 & 0.71 & 0.75 & 0.72 & 0.75 & 0.65 & 0.63 & 0.6 & 0.54 \\
attribute recognition & size & 0.72 & 0.77 & 0.76 & 0.83 & 0.8 & 0.64 & 0.8 & 0.81 & 0.56 & 0.63 & 0.68 & 0.68 \\
attribute recognition & texture & 0.8 & 0.77 & 0.78 & 0.82 & 0.8 & 0.79 & 0.71 & 0.77 & 0.65 & 0.63 & 0.66 & 0.67 \\
core object recognition & animals & 0.79 & 0.86 & 0.85 & 0.94 & 0.89 & 0.89 & 0.86 & 0.92 & 0.74 & 0.68 & 0.7 & 0.6 \\
core object recognition & artifacts & 0.8 & 0.83 & 0.8 & 0.85 & 0.81 & 0.76 & 0.78 & 0.78 & 0.72 & 0.77 & 0.83 & 0.72 \\
core object recognition & arts & 0.81 & 0.89 & 0.92 & 0.95 & 0.92 & 0.87 & 0.87 & 0.92 & 0.82 & 0.8 & 0.81 & 0.72 \\
core object recognition & clothing- accessories & 0.74 & 0.81 & 0.88 & 0.86 & 0.84 & 0.71 & 0.74 & 0.83 & 0.69 & 0.68 & 0.69 & 0.67 \\
core object recognition & common objects & 0.8 & 0.82 & 0.8 & 0.87 & 0.87 & 0.82 & 0.84 & 0.84 & 0.73 & 0.75 & 0.82 & 0.72 \\
core object recognition & food-beverage & 0.89 & 0.94 & 0.94 & 0.94 & 0.92 & 0.9 & 0.89 & 0.94 & 0.84 & 0.72 & 0.81 & 0.76 \\
core object recognition & indoor scenes & 0.75 & 0.89 & 0.88 & 0.87 & 0.88 & 0.83 & 0.88 & 0.85 & 0.78 & 0.85 & 0.88 & 0.72 \\
core object recognition & outdoor scenes & 0.88 & 0.88 & 0.94 & 0.91 & 0.91 & 0.84 & 0.89 & 0.89 & 0.77 & 0.85 & 0.9 & 0.82 \\
core object recognition & people & 0.76 & 0.87 & 0.87 & 0.86 & 0.83 & 0.78 & 0.77 & 0.82 & 0.69 & 0.81 & 0.72 & 0.69 \\
core object recognition & produce-plants & 0.9 & 0.93 & 0.9 & 0.94 & 0.92 & 0.85 & 0.86 & 0.95 & 0.87 & 0.79 & 0.76 & 0.87 \\
core object recognition & technology-electronics & 0.82 & 0.89 & 0.88 & 0.85 & 0.82 & 0.88 & 0.84 & 0.81 & 0.8 & 0.85 & 0.89 & 0.68 \\
core object recognition & vehicles & 0.68 & 0.85 & 0.9 & 0.86 & 0.78 & 0.75 & 0.77 & 0.82 & 0.74 & 0.86 & 0.89 & 0.64 \\
symbol recognition & app-tech icons & 0.77 & 0.81 & 0.83 & 0.83 & 0.83 & 0.85 & 0.83 & 0.81 & 0.79 & 0.73 & 0.77 & 0.73 \\
symbol recognition & astrological-zodiac signs & 0.74 & 0.77 & 0.85 & 0.89 & 0.89 & 0.98 & 0.82 & 0.82 & 0.92 & 0.56 & 0.69 & 0.61 \\
symbol recognition & currency symbols & 0.92 & 0.9 & 0.9 & 0.88 & 0.9 & 0.86 & 0.9 & 0.86 & 0.82 & 0.88 & 0.9 & 0.84 \\
symbol recognition & emoji & 0.7 & 0.78 & 0.78 & 0.78 & 0.74 & 0.67 & 0.78 & 0.81 & 0.62 & 0.74 & 0.7 & 0.67 \\
symbol recognition & flags & 0.64 & 0.75 & 0.83 & 0.8 & 0.79 & 0.71 & 0.75 & 0.8 & 0.64 & 0.58 & 0.73 & 0.57 \\
symbol recognition & formula & 0.66 & 0.65 & 0.73 & 0.68 & 0.76 & 0.52 & 0.71 & 0.68 & 0.53 & 0.63 & 0.68 & 0.53 \\
symbol recognition & logos-brands & 0.79 & 0.87 & 0.92 & 0.89 & 0.87 & 0.81 & 0.87 & 0.87 & 0.75 & 0.85 & 0.96 & 0.77 \\
symbol recognition & music & 0.84 & 0.85 & 0.91 & 0.89 & 0.89 & 0.89 & 0.82 & 0.82 & 0.81 & 0.84 & 0.8 & 0.81 \\
symbol recognition & religious symbols & 0.9 & 0.96 & 0.96 & 0.94 & 0.9 & 0.96 & 0.85 & 0.93 & 0.84 & 0.91 & 0.87 & 0.82 \\
symbol recognition & safety symbols & 0.85 & 0.96 & 0.95 & 0.92 & 0.94 & 0.88 & 0.9 & 0.94 & 0.83 & 0.92 & 0.92 & 0.81 \\
symbol recognition & traffic signs & 0.78 & 0.86 & 0.89 & 0.82 & 0.82 & 0.77 & 0.7 & 0.74 & 0.64 & 0.78 & 0.8 & 0.68 \\
text attribute recognition & books & 0.64 & 0.75 & 0.69 & 0.75 & 0.66 & 0.62 & 0.64 & 0.67 & 0.68 & 0.69 & 0.68 & 0.55 \\
text attribute recognition & documents & 0.78 & 0.83 & 0.86 & 0.91 & 0.89 & 0.8 & 0.85 & 0.88 & 0.76 & 0.73 & 0.73 & 0.68 \\
text attribute recognition & handwriting & 0.65 & 0.77 & 0.8 & 0.76 & 0.75 & 0.57 & 0.65 & 0.65 & 0.6 & 0.67 & 0.72 & 0.56 \\
text attribute recognition & lines & 0.66 & 0.67 & 0.72 & 0.75 & 0.71 & 0.78 & 0.67 & 0.76 & 0.65 & 0.57 & 0.61 & 0.61 \\
text attribute recognition & newsletter & 0.76 & 0.92 & 0.89 & 0.84 & 0.81 & 0.68 & 0.75 & 0.86 & 0.58 & 0.87 & 0.89 & 0.76 \\
text attribute recognition & number & 0.68 & 0.79 & 0.82 & 0.84 & 0.84 & 0.68 & 0.78 & 0.75 & 0.59 & 0.71 & 0.81 & 0.66 \\
text attribute recognition & power point slides & 0.78 & 0.81 & 0.81 & 0.86 & 0.84 & 0.71 & 0.75 & 0.83 & 0.65 & 0.7 & 0.75 & 0.64 \\
text attribute recognition & scene text & 0.73 & 0.85 & 0.89 & 0.93 & 0.9 & 0.79 & 0.75 & 0.86 & 0.66 & 0.78 & 0.77 & 0.63 \\
\bottomrule
\end{tabular}
}
\caption{Regex Based Extraction (RAE) Performance comparison across models for category: \textit{remember}}
\label{tab:remember_reg}
\end{table*}

\begin{table*}[t]
\centering
\small
\setlength{\tabcolsep}{4pt}
\resizebox{\textwidth}{!}{
\begin{tabular}{llcccccccccccccc}
\toprule
Sub-category & Sub-sub-category & \rotHeader{Gemma-3-4B-it\_en} & \rotHeader{Qwen2-VL-7B-Instruct\_en} & \rotHeader{Qwen2.5-VL-7B-Instruct\_en} & \rotHeader{Gemma-3-27b-it\_en} & \rotHeader{Gemma-3-12b-it\_en} & \rotHeader{GPT4omini\_en} & \rotHeader{Gemma-3-12B-it\_ar} & \rotHeader{Gemma-3-27B-it\_ar} & \rotHeader{GPT4omini\_ar} & \rotHeader{Qwen2-VL-7B-Instruct\_ar} & \rotHeader{Qwen2.5-VL-7B-Instruct\_ar} & \rotHeader{Gemma-3-4B-it\_ar}\\
\midrule
cognitive understanding & facial-emotional understanding & 0.93 & 0.99 & 0.93 & 0.93 & 0.89 & 0.89 & 0.85 & 0.89 & 0.85 & 0.86 & 0.86 & 0.89 \\
cognitive understanding & lingual expression alternation & 0.85 & 0.85 & 0.88 & 0.91 & 0.92 & 0.8 & 0.84 & 0.88 & 0.82 & 0.73 & 0.68 & 0.69 \\
cognitive understanding & semantic understanding (knowledge) & 0.95 & 0.98 & 0.99 & 0.94 & 0.97 & 0.97 & 0.98 & 0.97 & 0.93 & 0.93 & 0.97 & 0.92 \\
cognitive understanding & visual alternation & 0.9 & 0.93 & 0.94 & 0.97 & 0.94 & 0.96 & 0.96 & 0.97 & 0.93 & 0.87 & 0.9 & 0.85 \\
compositional attribute recognition & artistic style & 0.91 & 0.96 & 0.97 & 0.95 & 0.95 & 0.92 & 0.96 & 0.94 & 0.94 & 0.89 & 0.94 & 0.87 \\
compositional attribute recognition & shape & 0.83 & 0.84 & 0.8 & 0.84 & 0.88 & 0.78 & 0.85 & 0.78 & 0.68 & 0.74 & 0.72 & 0.72 \\
compositional attribute recognition & size & 0.67 & 0.79 & 0.81 & 0.86 & 0.91 & 0.74 & 0.83 & 0.84 & 0.73 & 0.61 & 0.7 & 0.59 \\
compositional attribute recognition & texture & 0.8 & 0.87 & 0.9 & 0.88 & 0.87 & 0.82 & 0.8 & 0.81 & 0.75 & 0.8 & 0.76 & 0.78 \\
compositional core object recognition & animals & 0.96 & 0.98 & 0.98 & 1.0 & 0.98 & 0.94 & 0.96 & 0.96 & 0.88 & 0.89 & 0.89 & 0.84 \\
compositional core object recognition & artifacts & 0.91 & 0.87 & 0.93 & 0.97 & 0.95 & 0.92 & 0.96 & 0.96 & 0.82 & 0.83 & 0.82 & 0.87 \\
compositional core object recognition & arts & 0.94 & 0.98 & 0.95 & 0.93 & 0.95 & 0.93 & 0.95 & 0.95 & 0.89 & 0.9 & 0.91 & 0.89 \\
compositional core object recognition & closed vocabulary object detection & 0.81 & 0.87 & 0.93 & 0.96 & 0.92 & 0.85 & 0.89 & 0.92 & 0.76 & 0.79 & 0.82 & 0.78 \\
compositional core object recognition & clothing-accessories & 0.89 & 0.95 & 0.99 & 1.0 & 0.94 & 0.96 & 0.92 & 0.94 & 0.88 & 0.85 & 0.9 & 0.79 \\
compositional core object recognition & common objects & 0.89 & 0.94 & 0.89 & 0.97 & 0.95 & 0.89 & 0.9 & 0.9 & 0.84 & 0.89 & 0.9 & 0.82 \\
compositional core object recognition & food-beverage & 0.94 & 0.98 & 0.99 & 0.96 & 0.94 & 0.91 & 0.96 & 0.95 & 0.88 & 0.92 & 0.89 & 0.92 \\
compositional core object recognition & indoor scenes & 0.95 & 1.0 & 0.98 & 1.0 & 1.0 & 0.95 & 0.98 & 1.0 & 0.92 & 0.97 & 0.98 & 0.92 \\
compositional core object recognition & outdoor scenes & 0.95 & 0.96 & 0.98 & 0.96 & 0.98 & 0.98 & 0.99 & 0.98 & 1.0 & 0.96 & 0.94 & 0.94 \\
compositional core object recognition & people & 0.93 & 0.99 & 0.97 & 0.96 & 0.97 & 0.96 & 0.93 & 0.99 & 0.93 & 0.95 & 0.95 & 0.95 \\
compositional core object recognition & produce-plants & 0.88 & 0.94 & 0.93 & 0.93 & 0.96 & 0.97 & 0.93 & 0.94 & 0.88 & 0.84 & 0.91 & 0.88 \\
compositional core object recognition & technology-electronics & 0.95 & 0.96 & 0.95 & 0.96 & 0.98 & 0.87 & 0.95 & 0.96 & 0.89 & 0.91 & 0.93 & 0.87 \\
compositional core object recognition & vehicles & 0.91 & 0.96 & 0.97 & 0.96 & 1.0 & 0.96 & 0.96 & 0.96 & 0.89 & 0.95 & 0.96 & 0.89 \\
\bottomrule
\end{tabular}
}
\caption{Regex Based Extraction (RAE) Performance comparison across models for category: \textit{understand}}
\label{tab:understand_reg}
\end{table*}

% \section{}
% \begin{table}[h]
% \centering
% \small
% \begin{tabular}{lcccc}
% \toprule
% \textbf{Cognitive Level} & \textbf{Sample Size} & \textbf{Valid (Agree)} & \textbf{Flagged} & \textbf{Quality Rate} \
% \midrule
% Remember & 361 & 354 & 6 & 98.1% \
% Understand & 202 & 201 & 1 & 99.5% \
% Apply & 53 & 51 & 0 & 96.2% \
% Analyze & 168 & 162 & 4 & 96.4% \
% Evaluate & 85 & 82 & 2 & 96.5% \
% Create & 100 & 94 & 2 & 94.0% \
% \midrule
% \textbf{Total / Average} & \textbf{969} & \textbf{944} & \textbf{15} & \textbf{97.4%} \
% \bottomrule
% \end{tabular}
% \caption{\textbf{Validation Statistics.} Quality assurance results on a stratified subset ($N=969$) evaluated by Gemini 3 Pro and audited by humans.}
% \label{tab:validation}
% \end{table}
\end{document}